\documentclass[nohyperref]{article}

\usepackage{microtype}
\usepackage{CJKutf8}
\usepackage{graphicx}
\usepackage{wrapfig}
\usepackage{tabularx}
\usepackage{subfigure}
\usepackage{dsfont}
\usepackage{booktabs} % for professional tables
\newcommand\numberthis{\addtocounter{equation}{1}\tag{\theequation}}

\usepackage{hyperref}

\usepackage[accepted]{icml2022}

\usepackage{amsmath}
\usepackage{amssymb}
\usepackage{mathtools}
\usepackage{amsthm}

\usepackage[capitalize,noabbrev]{cleveref}

\theoremstyle{plain}
\newtheorem{theorem}{Theorem}[section]

\theoremstyle{definition}
\newtheorem{definition}[theorem]{Definition}

\theoremstyle{remark}

\usepackage[textsize=tiny]{todonotes}

\icmltitlerunning{Generalized Beliefs for Cooperative AI}

\begin{document}

\definecolor{luisas_color}{HTML}{3e9555}
\definecolor{shimons_color}{HTML}{fe8655}
\definecolor{darius_color}{HTML}{9d8655}
\definecolor{jakobs_color}{HTML}{fe8355}
\definecolor{christians_color}{HTML}{793dae}
\newcommand{\luisa}[1]{{\color{luisas_color}{LZ: #1}}}
\newcommand{\chris}[1]{{\color{christians_color}}{CS: #1}}
\newcommand{\darius}[1]{{\color{darius_color}{DM: #1}}}
\newcommand{\sw}[1]{{\color{shimons_color}{SW: #1}}}
\newcommand{\jakob}[1]{{\color{jakobs_color}}{JF: #1}}

\twocolumn[

\icmltitle{Generalized Beliefs for Cooperative AI}




\begin{icmlauthorlist}
\icmlauthor{Darius Muglich}{yyy}
\icmlauthor{Luisa Zintgraf}{yyy}
\icmlauthor{Christian Schroeder de Witt}{yyy}
\icmlauthor{Shimon Whiteson}{yyy}
\icmlauthor{Jakob Foerster}{yyy}

\end{icmlauthorlist}

\icmlaffiliation{yyy}{University of Oxford, England, United Kingdom}
%\icmlaffiliation{comp}{Company Name, Location, Country}
%\icmlaffiliation{sch}{School of ZZZ, Institute of WWW, Location, Country}

\icmlcorrespondingauthor{Darius Muglich}{dariusm1997@yahoo.com}

\icmlkeywords{Machine Learning, ICML}

\vskip 0.3in
]



\printAffiliationsAndNotice{} % otherwise use the standard text.

\begin{abstract}
Self-play is a common paradigm for constructing solutions in Markov games that can yield optimal policies in collaborative settings. However, these policies often adopt highly-specialized conventions that make playing with a novel partner difficult. To address this, recent approaches rely on encoding symmetry and convention-awareness into policy training, but these require strong environmental assumptions and can complicate policy training. We therefore propose moving the learning of conventions to the belief space. Specifically, we propose a belief learning model that can maintain beliefs over rollouts of policies not seen at training time, and can thus decode and adapt to novel conventions at test time. We show how to leverage this model for both search and training of a best response over various pools of policies to greatly improve ad-hoc teamplay. We also show how our setup promotes explainability and interpretability of nuanced agent conventions.
\end{abstract}

\section{Introduction}
\label{sec:intro}

One of the most popular ways of training a policy in multi-agent reinforcement learning (MARL) is via \emph{self-play} \cite{samuel1959some}, where during training a joint policy controls the strategies of every player. This can lead to convergence to highly specialized policies that have enjoyed great success in zero-sum settings \cite{brown2020combining, silver2018general}. Self-play can also yield highly proficient policies in cooperative settings \cite{hu2019simplified, schroeder2019multi, yu2021surprising}; however, if self-play is left unconstrained, agents that train independently with self-play and then meet at test time to play a cooperative game (i.e., engage in \emph{cross-play}) may do much worse than when they were self-playing.

Agents in cooperative tasks often must learn to implicitly communicate, particularly when the available communication channels are too narrow to fully resolve mutual uncertainty \citep{heider1944experimental, tian2020learning}. In the case of the popular and frequently-studied card game Hanabi \citep{bard2020hanabi} (see Appendix \ref{appendix:hanabi-rules} for details on Hanabi), consider the example \textit{convention} that ``hinting green'' also means that the rightmost held card of the player receiving the hint is playable. One could then construct a second policy where instead ``hinting yellow'' bears this implicit cue. While these two policies are essentially equivalent, if two agents meet and one plays the former policy and the other plays the latter, they will perform poorly because they have each overfit to their respective conventions.

To improve the cross-play performance of independent self-play trained agents (i.e., to effectively engage in \textit{zero-shot coordination} \cite{hu2020other}), recent approaches have explored explicitly encoding environmental symmetries (where symmetry refers to state/action sequences that are equivalent up to relabelling of the state/action space) into policy training \cite{hu2020other}. However, this complicates the training process and requires domain knowledge. 

Others have explored modular approaches that separately learn the complexities intrinsic to the task (i.e., rule-dependent behaviour) and the conventions specific to a partner (i.e., convention-dependent behaviour) \cite{shih2021critical}. However, their method comes at the expense of scalability, with experiments limited to bandit problems and 1-colour Hanabi. 

Yet another approach that has been explored is in controlling the cognitive-reasoning depth (i.e. controlling the extent to which agents build more complex conventions on top of the baseline expectation that earlier conventions are being followed) so to avoid formation of arbitrary conventions that can complicate cross-play \citep{hu2021off}. While an interesting direction, it mitigates policies from adopting a large class of behaviours, and as well motivates the question of whether the true absence of such convention formation is strictly necessary for coordination?

We seek a paradigm to learn policies without the disadvantages listed above; that is, we seek to relax strong environmental assumptions, retain scalability, and to leave policy learning unconstrained. As such, in this paper, we propose shifting the burden of learning conventions and intent onto the agent's maintained model of uncertainty over the environment, i.e., its \emph{belief}. Specifically, we make the following main contributions:

\begin{enumerate}
    \item We propose a methodology to learn an emulated belief (bypassing costly exact Bayesian updating) that can be used to decode agent intent and behaviour, and test this methodology on the \textit{generalized} belief modelling task, i.e., the zero-shot task of maintaining belief over trajectories featuring a policy not seen at training time (this is a sequence-modelling task, where the trajectory is encoded and the belief state is decoded). %This automates the learning of symmetry in the Dec-POMDP rather than explicitly encoding it as in \citet{hu2020other}, so that less prior knowledge on the Dec-POMDP is assumed.
    
    \item We leverage this emulated belief for improving cross-play performance via techniques such as Monte Carlo search and training a best response over a pool of policies. In this way we propose effective auxiliary mechanisms that can be flexibly added to any policy regime to enhance cross-play, whilst keeping training simple and scalable.
    
    \item We use our belief model to promote interpretability of agent conventions by giving evidence to how turns of a game interrelate, and what environmental features agents use for implicit communication. Interpretability is critical for building trustworthy and safe systems \cite{wells2021explainable}.
\end{enumerate}

\section{Background}

This section formalises the problem setting and defines the notion of beliefs.

\subsection{Dec-POMDPs}\label{sec:dec-pomdp}

We formalise the cooperative multi-agent setting as a decentralised partially-observable Markov decision process \citep[Dec-POMDP]{oliehoek2012decentralized} which is a 9-tuple $(\mathcal{S},$ $\mathcal{N},$ $\{\mathcal{A}^i\}_{i=1}^n,$ $\{\mathcal{O}^i\}_{i=1}^n,$ $\mathcal{T},$ $\mathcal{R},$ $\{\mathcal{U}^i\}_{i=1}^n,$ $T,$ $\gamma$), for finite sets $\mathcal{S}, \mathcal{N},$ $\{\mathcal{A}^i\}_{i=1}^n,$ $\{\mathcal{O}^i\}_{i=1}^n$, denoting the set of states, agents, actions and observations, respectively, where a superscript $i$ denotes the set pertaining to agent $i \in \mathcal{N}=\{1 , \dots, n\}$ (i.e., $\mathcal{A}^i$ and $\mathcal{O}^i$ are the action and observation sets for agent $i$, and $a^i \in \mathcal{A}^i$ and $o^i \in \mathcal{O}^i$ are a specific action and observation agent $i$ may undertake). We also write $\mathcal{A}=\times_i \mathcal{A}^i$ and $\mathcal{O}=\times_i \mathcal{O}^i$, the sets of joint actions and observations, respectively. $s_t \in \mathcal{S}$ is the state at time $t$ and $s_t=\{s^k_t\}_k$, where $s_t^k$ is state feature $k$ of $s_t$. %(state features refer to the finitely-many enumerated characteristics that together specify the environment state)
$a_t \in \mathcal{A}$ is the joint action of all agents taken at time $t$, which  changes the state according to the transition distribution $s_{t+1} \sim \mathcal{T}(s_{t+1} \ | \ s_t, a_t)$. The subsequent joint observation of the agents is $o_{t+1} \in \mathcal{O}$, distributed according to $o_{t+1} \sim \mathcal{U}(o_{t+1} \ | \ s_{t+1}, a_t)$, where $\mathcal{U}=\times_i \mathcal{U}^i$; observation features of $o_{t+1}^i \in \mathcal{O}^i$ are notated analogously to state features; that is, $o_{t+1}^i=\{o_{t+1}^{i,k}\}_k$. The reward $r_{t+1} \in \mathbb{R}$ is distributed according to $r_{t+1} \sim \mathcal{R}(r_{t+1} \ | \ s_{t+1}, a_t)$. $T$ is the horizon and $\gamma \in [0,1]$ is the discount factor.

Notating $\tau_t^i=(a_0^i,o_1^i,\dots,a_{t-1}^i,o_t^i)$ for the action-observation history of agent $i$, agent $i$ acts according to a policy $a_t^i \sim \pi^i(a_t^i \ | \ \tau_t^i)$. The agents seek to maximize the return, i.e., the expected discounted sum of rewards:
\begin{equation}
J_T := \mathbb{E}_{p(\tau_T)}[\sum_{t'\leq T} \gamma^{t'-1} r_{t'}],
\end{equation}
where $\tau_t = (s_0, a_0, o_1, r_1, \dots, a_{t-1}, o_t, r_t, s_t)$
is the trajectory until time $t$.

\subsection{Beliefs}\label{sec:beliefs}

As the Dec-POMDP is not fully observable, an agent $i$ can operate under an estimate of the trajectory. The private belief of agent $i$ is  the posterior $b_t^i := p(\tau_t \ | \ \tau_t^i )$, i.e., the model of uncertainty agent $i$ maintains about the trajectory and true state. The belief is a sufficient statistic for the environment state and characterizes the theory of mind \cite{baker2017rational} often requisite for successful performance in Dec-POMDPs.

The private belief in a Dec-POMDP may be iteratively maintained if the policy of agent $j$ is prior knowledge. If so, then each action taken by agent $j$ introduces a belief update across all other agents: supposing agent $i$ has current belief $b_{t-1}^i$ and next observes $(a_t^j, o_t^i)$ (where we have broken out the action of player $j$ from the observation), we can use Bayes' rule to obtain
\begin{align*}\label{eq:dec-pomdp-belief-update}
    b_t^i&=p(\tau_t \ | \ \tau_t^i)
    \\&=p(\tau_t \ | \ \tau_{t-1}^i, a_t^j, o_t^i)
    \\&=\frac{b_{t-1}^i\pi^j(a_t^j \ | \ \tau_{t-1})p(o_t^i \ | \ \tau_{t-1}, a_t^j)}{\sum_{\tau'_{t-1}}b_{t-1}^i\pi^j(a_t^j \ | \ \tau_{t-1}')p(o_t^i \ | \ \tau_{t-1}',a_t^j)}. \numberthis
\end{align*}
However, this manner of belief updating not only assumes $\pi^j$ is prior knowledge, but also requires evaluation over all possible trajectories and is thus computationally intractable in large settings. To circumvent these issues, an aggregator function such as a recurrent neural network is commonly relied on to take the action-observation history so for its intermediate hidden states to form a statistic $z_t^i$ of the history sufficient for predicting future observations \cite{hausknecht2015deep, zhang2015policy, zhu2017improving}. Often $z_t^i$ is trained by conditioning the policy on it and using gradient descent. However, the RL signal is often too weak to learn a rich representation for $z_t^i$ that provides sufficient statistics for the filtering posterior over states, a phenomenon that has been empirically demonstrated, e.g., by \citet{moreno2018neural} and \citet{zintgraf2020varibad}. We therefore look neither to maintain explicit Bayesian updates nor to implicitly form sufficient statistics, but to \textit{learn} a belief emulation.

\section{Model}\label{sec:model}

For generalized belief learning, our approach is to learn a belief model using supervised learning over rollouts of several different self-play trained policies, and our main question is whether this is sufficient to generalize reasoning over novel policies at test time. See Appendix \ref{appendix:autoregressive-beliefs} for further specification of learning beliefs in the Dec-POMDP setting and our autoregressive approach to belief modelling.

We largely use the same self-attention based architecture as \citet{vaswani2017attention}, but we modify the embedding mechanisms, which we describe presently.

Without loss of generality, we fix an agent $i$. Let $H = |\{c_1, \dots, c_H\}|$ be the number of unobservable state features\footnote{If $H$ varies per timestep, set $H=\max\{H_1,\dots,H_T\}$.}. To input the action-observation history and the conditioned unobserved features, we require learnable embedding functions $\Psi_E : (\mathcal{O}^i)^T \mapsto \mathbb{R}^{T \times d}, \Psi_D : \mathcal{S}^H \mapsto \mathbb{R}^{H \times d}$ such that
\begin{align*}
    \Psi_E(\tau_t^i) = \textbf{x}_E, \text{  }\Psi_D(c_1,\dots,c_H)=\textbf{x}_D,
\end{align*}
where $\textbf{x}_E, \textbf{x}_D$ are then inputted to the encoder and decoder modules of our architecture, respectively \cite{vaswani2017attention}. Here $T$ and $d$ are the maximum timestep\footnote{If $t<T$, in practice we pad $\tau_t^i$ and $\textbf{x}_E$ to make dimensions match.} and the dimensionality of the embedding, respectively, where both are set as hyperparameters.

\begin{figure}
    \centering
    \includegraphics[width=80mm]{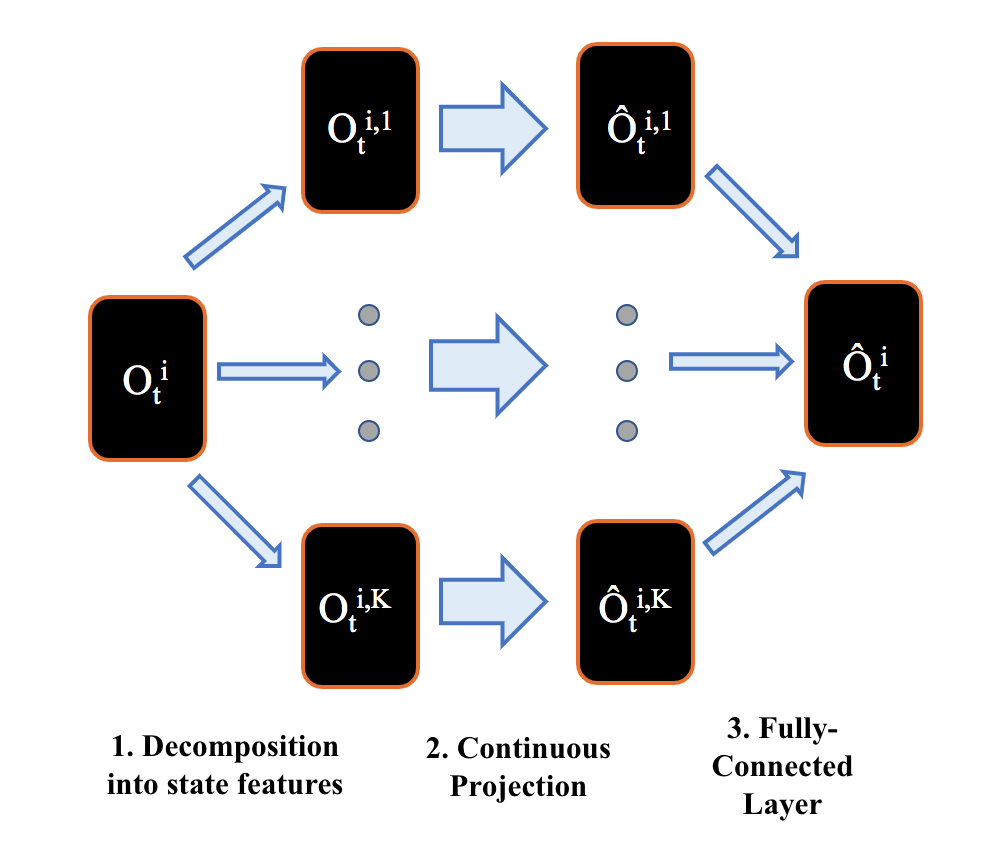}
    \caption{Proposed input embedding scheme for encoder module ($\Psi_E$), where we decompose a state into its features, project the features onto an information space, then aggregate them to obtain the state embedding.}
    \label{fig:embedding-func}
\end{figure}

For $\Psi_E$, we apply a trainable embedding layer over the discrete observable state features $o_t^{i,1},\dots,o_t^{i,K},$ where $K=|o_t^i|$ for all $t$, so as to project them onto a continuous vector space of dimension $d_\text{feature}$ to obtain $\hat{o}_t^{i,1},\dots,\hat{o}_t^{i,K}$. These embedding representations of the observable state features then become parameters of our model and are learned through the contexts the state features appear in during gameplay.\\

Following this continuous projection of the state features, a fully connected layer is applied over them so as to obtain an embedding $\hat{o}_t^i$ of the observable state $o_t^i$. The representations $\hat{o}_t^i$ for each $t$ comprise $\textbf{x}_E$ and are passed to the learning system. 
\looseness=-1

For optimisation purposes, we add dropout to $\Psi_E$ for regularisation \citep{srivastava2014dropout}, layer normalisation to assist with learning \citep{ba2016layer}, and a positional embedding to provide a learnable signal encoding the timestep order \citep{vaswani2017attention}. Figure \ref{fig:embedding-func} provides a high-level illustration of $\Psi_E$. For $\Psi_D$, we maintain an embedding layer of size $H$ to project $c_1,\dots,c_H$ onto a continuous space.
\looseness=-1

Our choice of $\Psi_E$ does not directly project $o_t^i$ onto a continuous space to obtain $\hat{o}_t^i$ because the continuous projective layer must maintain an enumeration of all members in the vocabulary. Letting $d_k$ denote the number of values that observable state feature $o_t^{i,k}$ may take on, we have that the vocabulary for all $o_t^i$ is of size $\prod_{k=1}^Kd_k$. 

Alternatively, using our choice of $\Psi_E$, the vocabulary we need maintain (over the state features) is just of size $\sum_{k=1}^K d_k$. Besides memory considerations, training on large vocabularies can be difficult and can result in poor performance on rarer tokens \citep{labeau2018neural}.

Furthermore, the reason we combine the continuous state feature representations $\hat{o}_t^{i,1},\dots,\hat{o}_t^{i,K}$ in a fully-connected layer to obtain $\hat{o}_t^i$ is two-fold: 

\begin{enumerate}
\item Without this fully-connected layer, $\textbf{x}_E$ would be of size $KT \times d$ rather than $T \times d$, which is problematic given the quadratic memory and computation requirement of the dot-product self-attention mechanism: the fully-connected layer thus reduces the memory requirement from $O(K^2T^2)$ to $O(T^2)$ and the compute requirement from $O(dK^2T^2)$ to $O(dT^2),$ which on the Hanabi control task we have found to be the difference between prohibitively costly and efficiently executable; 

\item The fully-connected layer gives context that all the state features belong to the same state, providing the model with meaningful structure about the data; this context can in turn aid with the problem of poor performance on rarer states, as it gives the model a means to compare states with similar state feature values.

\end{enumerate}

%In the next three sections, we introduce the generalized belief model (Section \ref{sec:generalized-beliefs}), and demonstrate its tripartite ability to 1) convincingly work as a model of uncertainty over rollouts of a novel policy (Section \ref{sec:generalized-beliefs}); 2) leverage its capacity for convention generalization to improve cross-play (Section \ref{sec:improving-xp}); 3) promote explainability and give insight as to what channels a black-box policy uses to convey implicit information (Section \ref{sec:belief-introspec}).

% \clearpage

\section{Generalized Belief Learning}\label{sec:generalized-beliefs}

The training of the belief model is set up so as to be similar to policy learning, rather than the typical supervised regime of fixing a training set, a validation set and a test set: given a collection of pre-trained policies, we run a number of parallel simulators to simultaneously record the trajectories of games played by the pre-trained policies to an experience replay buffer. 

Concurrently, we sample from the experience replay buffer and use the sampled rollouts for training the belief model with a cross entropy loss (see Appendix \ref{appendix:autoregressive-beliefs}). Training in such a way helps avoid over-fitting without manually tuning hyper-parameters and regularization. See Appendix \ref{appendix:belief-learning} for the hyperparameters used.

Section \ref{sec:experimental-setup} describes the experimental setup used, and Section \ref{sec:results} shows the findings of the experiments.

\subsection{Experimental Setup}\label{sec:experimental-setup}

We use the AI benchmark task and representative Dec-POMDP Hanabi \cite{bard2020hanabi} for our experiments. Hanabi is a unique and challenging card game that requires agents to formulate informative implicit conventions in order to be successful (see Appendix \ref{appendix:hanabi-rules}).

We used thirteen pre-trained simplified action decoder (SAD) policies that were used in the work of \citet{hu2020other}, and which we downloaded from their GitHub repository.\footnote{\url{https://github.com/facebookresearch/hanabi_SAD}} 
In addition, we trained 12 Other-Play (OP) policies \citep{hu2020other}. 
OP is a training regime for producing a class of policies that can cohesively maintain environmental symmetry so to effectively engage in zero-shot coordination. SAD and OP policies adopt markedly different styles of play and use very different conventions (as we shall explore more in depth), and so their complementary analysis will evince how our proposed methodologies can work for varied policy types. The OP policies were trained with SAD and Value Decomposition Networks \cite{sunehag2017value}. The policies in each pool differ by the seeds used for training, which were randomly chosen for each model.

The policies achieve average scores of $23.97 \pm 0.04$ in two-player Hanabi self-play (a perfect score in Hanabi is 25), but only average scores of $6.258 \pm 1.51$ in cross-play. 
This highlights just how specialized the respective conventions adopted in these policies are, where despite all the policies being trained with the same method and playing to analogous levels of proficiency, they lose much of their performance capability in cross-play. The OP policies used achieve average self-play scores of $23.1 \pm 0.05$, but cross-play scores of $16.88 \pm 0.96$.
\looseness=-1

We henceforth refer to a belief model trained with trajectories featuring only one policy as a ``Single'' model, and a model trained with trajectories featuring multiple polices as a ``Multi'' model (a Multi model constitutes a ``generalized'' belief). ``Multi''-$x$ will refer to a Multi model trained with trajectories of $x$ different policies.

Over both our SAD and OP pools, we trained several belief models: for the SAD pool, we randomly selected various policies to train 8 different Single beliefs, 6 different Multi-6 models, and 6 different Multi-12 models; for the OP pool, we similarly trained 8 different Single beliefs, 6 different Multi-6 models, and 6 different Multi-12 models. We then tested how these various models fared with maintaining belief over the policies they did not see at training time.

All the models were otherwise trained with the procedure described at the beginning of Section \ref{sec:generalized-beliefs}. Everything but the trajectories trained with were fixed so as to mitigate the effect of confounding variables. The trained models were then tested against trajectories featuring the held out policy. See Appendix \ref{appendix:belief-learning} for more details.

\subsection{Grounded Belief}\label{sec:grounded-belief}
We compare against the grounded belief $\psi_0$ for agent $i$:
\begin{equation}
    \psi_0(\tau_t \ | \ \tau_t^i) = \prod_{h=1}^H\mathds{1}_{x=c_h} f(x=c_h \ | \ \tau_t^i),
\end{equation}
where $f$ is the probability mass function for the categorical distribution over the plausible values the unobservable state features may take on, where plausibility is determined by conditioning on \textit{grounded} information. Grounded information is information that may be derived from the action-observation history only as can be deduced from the rules of the Dec-POMDP, and not from assumptions of what a policy may intend from certain behaviours. As such, the grounded belief assumes knowledge of environmental dynamics, but it does not take into account implicit cooperative cues that may be immanent of agent play. It is thus an important standard for testing against so as to measure how much implicit information our methodology takes into account.

\subsection{Results}\label{sec:results}

% \begin{figure}
%     \centering
%     \begin{minipage}{1.15\linewidth}
%     \hspace{-0.4cm}\includegraphics[width=1.0\linewidth]{generalized-belief-comparison.png}
%     \end{minipage}
%     \vspace{-1.0\baselineskip}
%     \caption{Per card cross entropy scores of the average of the Single models and the different Multi models, all tasked with maintaining belief over trajectories featuring a policy not seen at training time. The grounded belief and the belief model trained on the test policy itself are provided here for reference. The shading corresponds to the standard error of the mean at each timestep. The curves were computed over 20k randomly generated games.}
%     \label{fig:belief-curves}
% \end{figure}
\begin{figure*}[hbt!]
    \centering
    \begin{minipage}{0.48\linewidth}
    \includegraphics[width=1.0\linewidth]{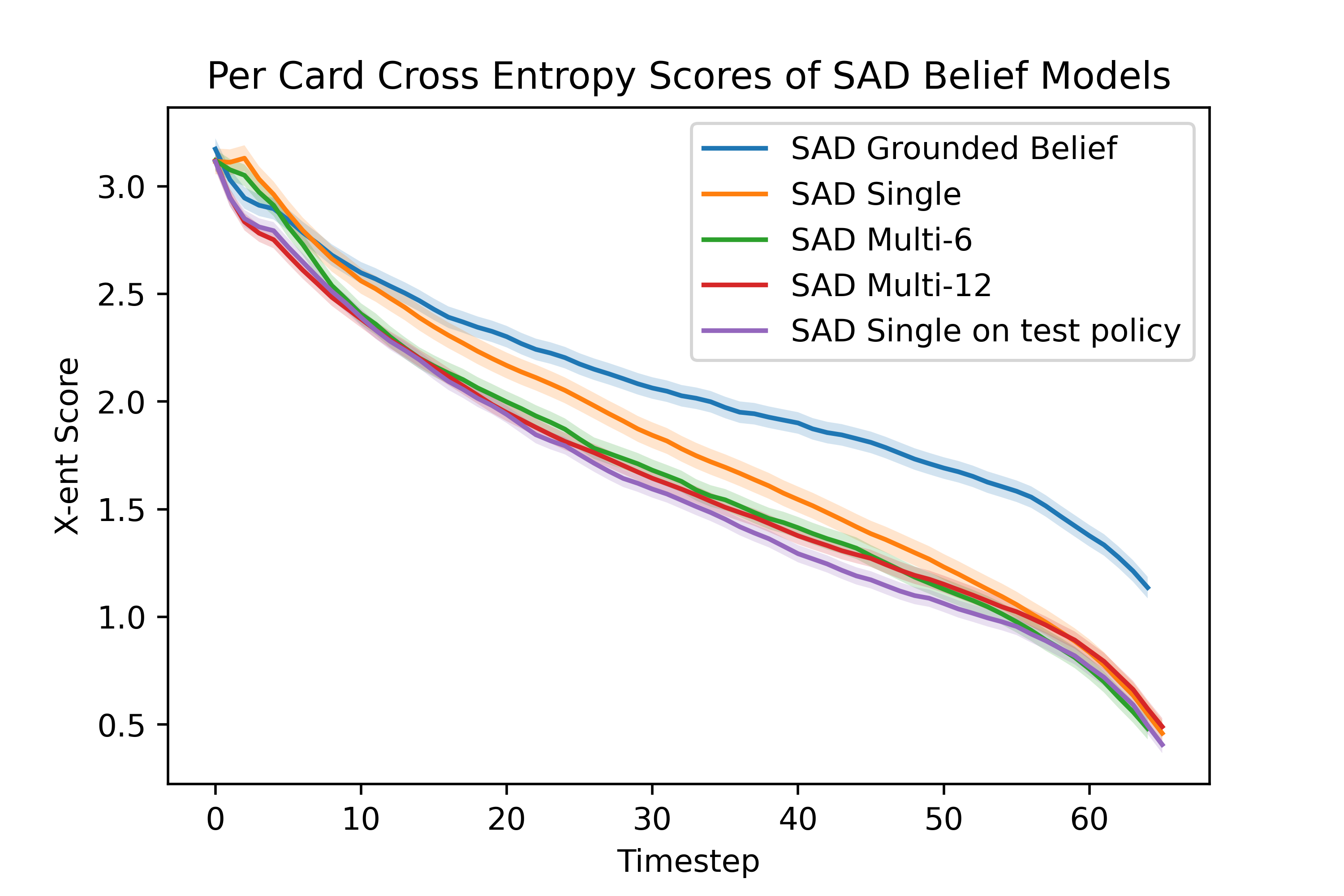}
    \end{minipage}
    \begin{minipage}{0.48\linewidth}
    \includegraphics[width=1.0\linewidth]{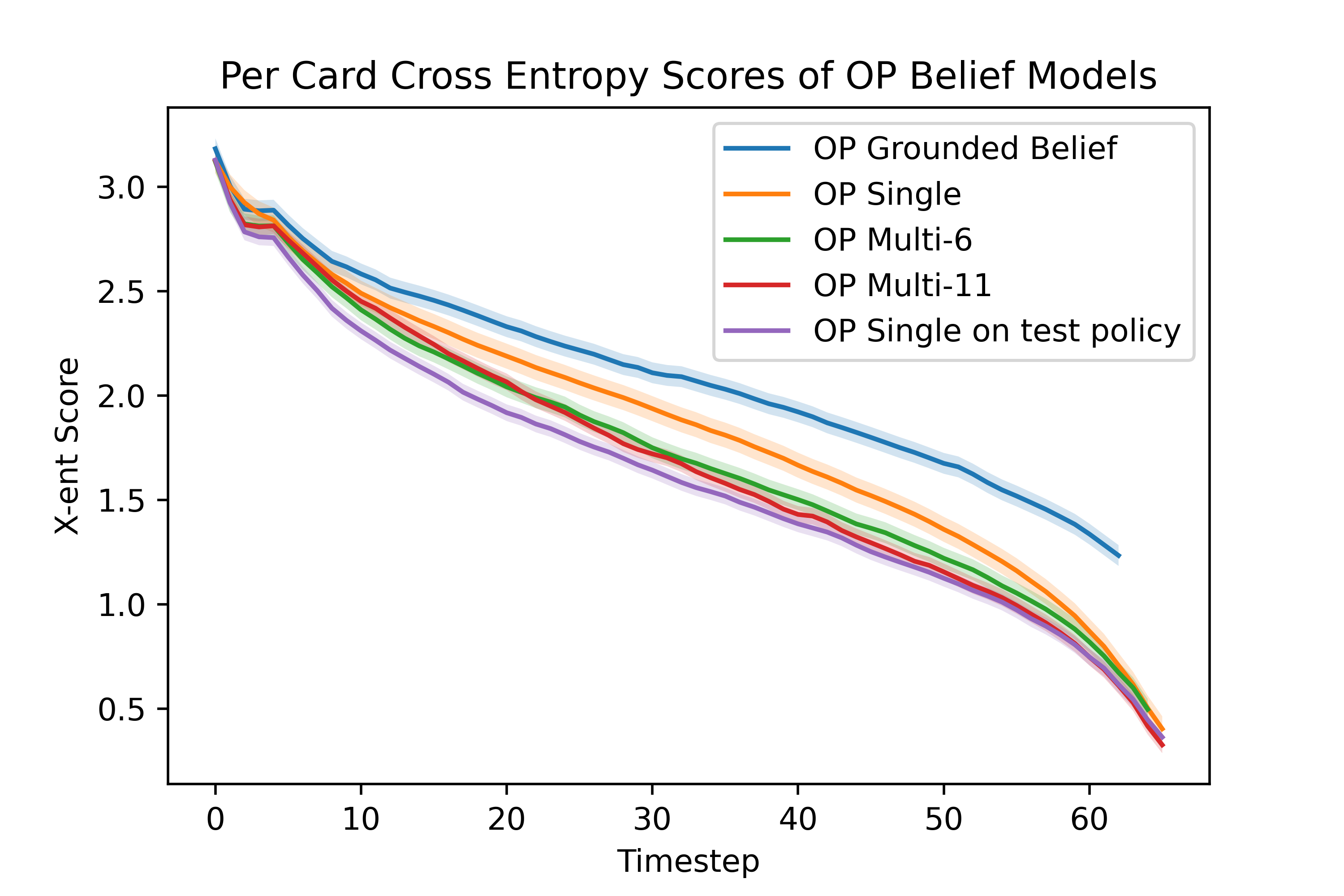}
    \end{minipage}
    \begin{minipage}{0.48\linewidth}
    \includegraphics[width=1.0\linewidth]{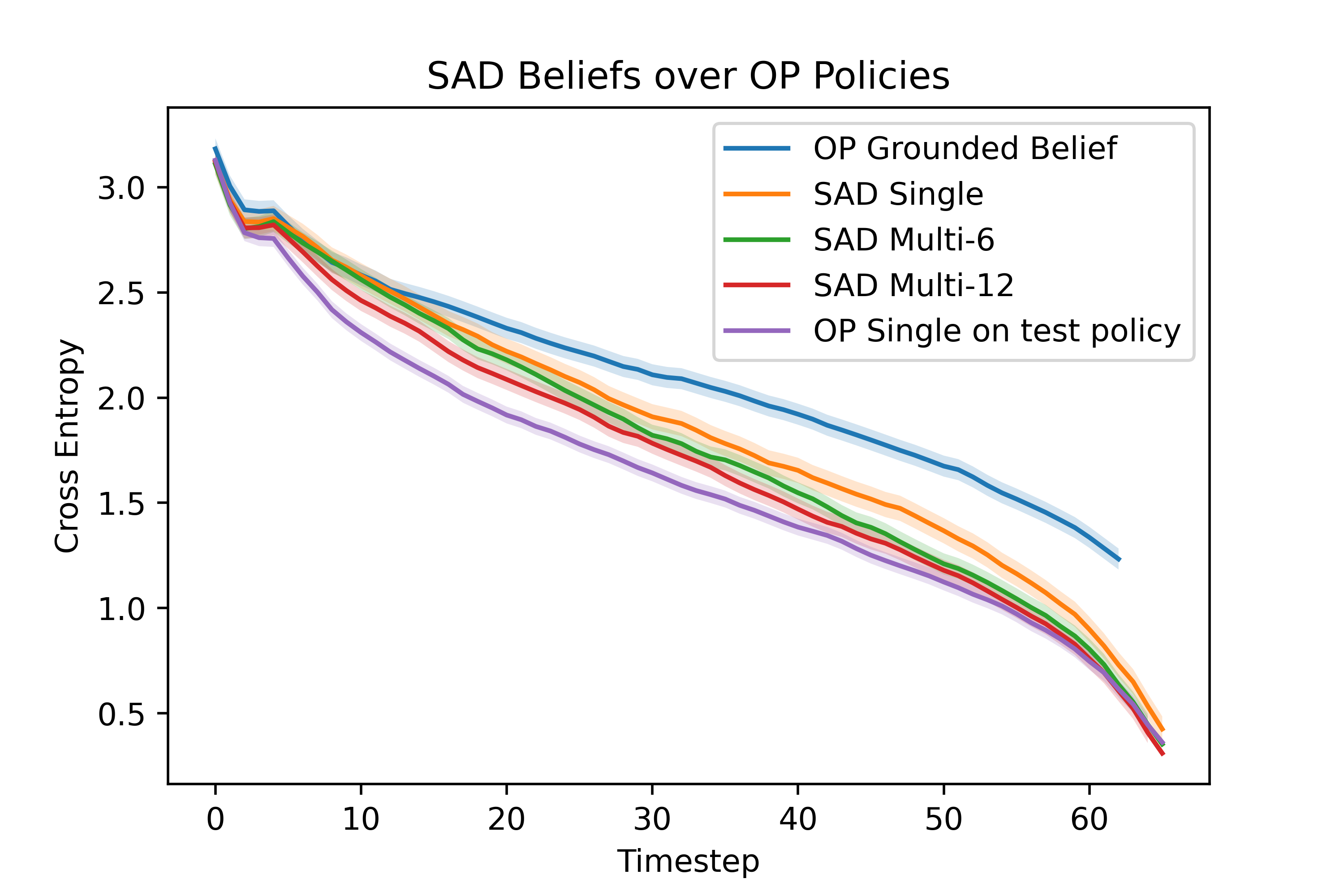}
    \end{minipage}
    \caption{Per card cross entropy (X-ent) scores of the averages of the Single models and the different Multi models, all tasked with maintaining belief over trajectories featuring policies not seen at training time. The grounded belief and the belief model trained on the test policy itself are provided here for reference. The shading corresponds to the standard error of the mean at each timestep. The curves were computed over 20k randomly generated games.}
    \label{fig:belief-curves}
\end{figure*}

\begin{figure*}[hbt!]
    \centering
    \begin{minipage}{0.48\linewidth}
    \includegraphics[width=1.0\linewidth]{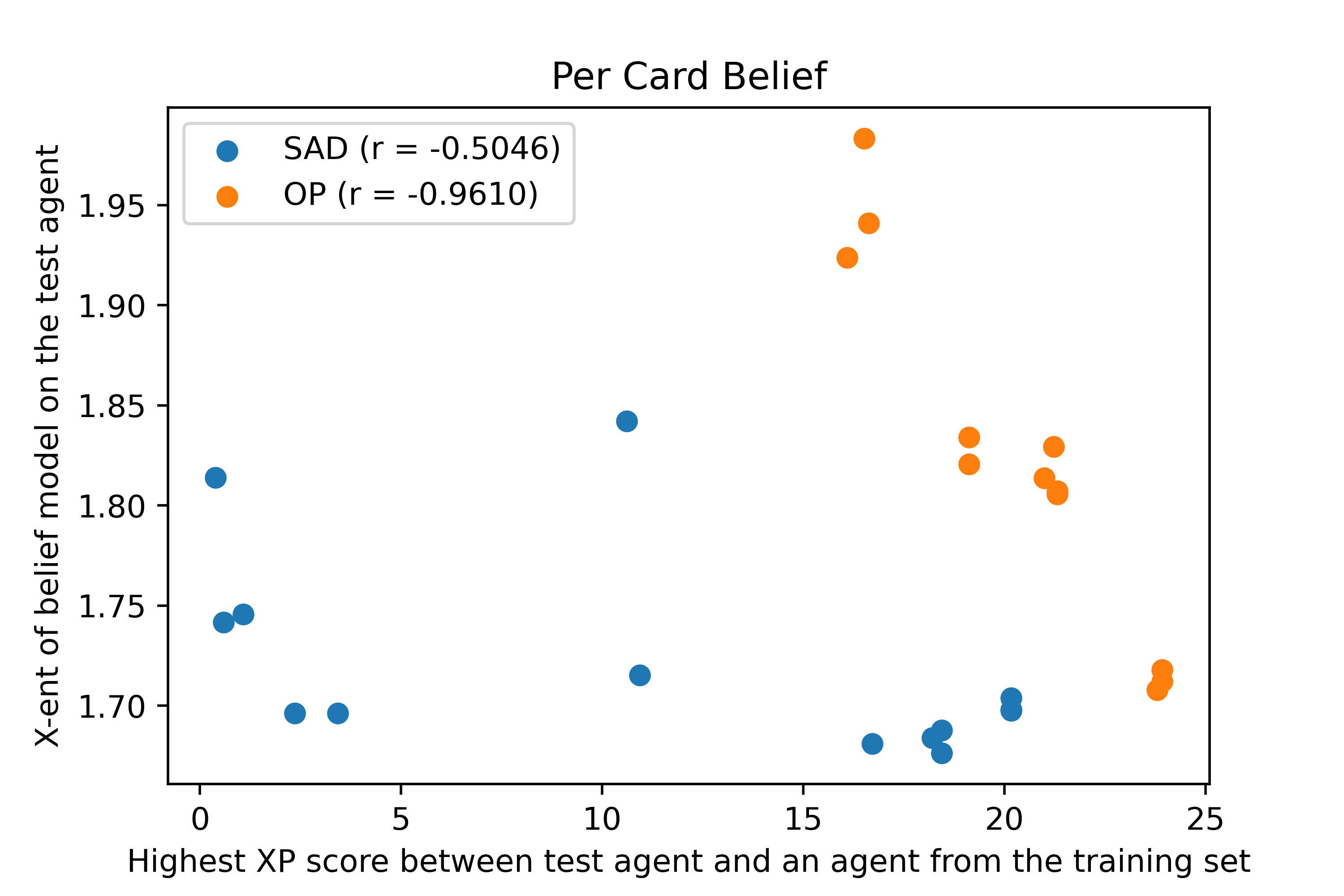}
    \end{minipage}
    \begin{minipage}{0.48\linewidth}
    \includegraphics[width=1.0\linewidth]{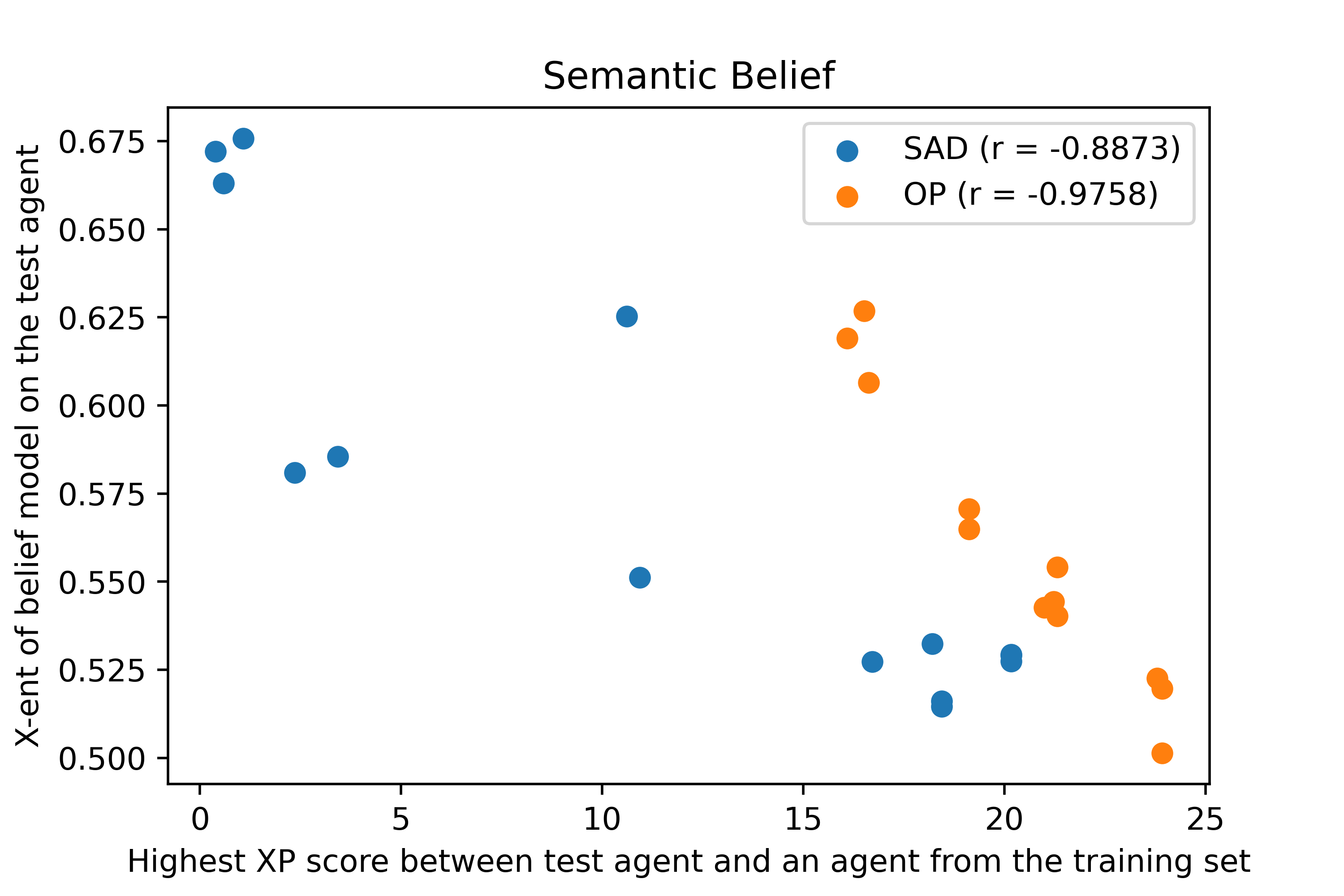}
    \end{minipage}
    \caption{Left: Tested on two Multi-6 per card belief models. Right: Tested on two Multi-6 semantic belief models. The vertical axis represents the average cross entropy of the Multi-6 over a test policy, and the horizontal axis represents the highest cross-play (XP) score between the test agent and an agent from the training set. $r$ denotes the Pearson's correlation coefficient \citep{benesty2009pearson}.}
    \label{fig:per-card-semantic-scatter}
\end{figure*}

%\begin{figure}[hbt!]
%    \centering
%    \includegraphics[width=70mm]{per-card-belief-scatter.png}
%    \caption{Tested on two Multi-6 per card belief models, the vertical axis represents the average cross entropy of the Multi-6 over a test policy, and the horizontal axis represents the highest cross-play score between the test agent and an agent from the training set.}
%    \label{fig:per-card-scatter}
%\end{figure}
%\begin{figure}[hbt!]  
%    \centering
%    \includegraphics[width=70mm]{semantic-belief-scatter.png}
%    \caption{Tested on two Multi-6 semantic belief models, the vertical axis represents the average cross entropy of the Multi-6 over a test policy, and the horizontal axis represents the highest cross-play score between the test agent and an agent from the training set.}
%    \label{fig:semantic-scatter}
%\end{figure}

Figure \ref{fig:belief-curves} shows the results of our generalized belief learning methodology (see Appendix \ref{appendix:ce-interpretation} for an intuitive interpretation of cross entropy scores over unobservable features in a Dec-POMDP). The Multi models can predict over rollouts of the unseen test policy with nearly the accuracy of the model trained on the test policy itself.

One may suspect there to be a correlation between how ``aligned'' the training policies are with the test policy, and the belief model's performance over the test policy. As such, to discuss the correlation between cross-play and cross entropy, we consider the \textit{semantic} belief: rather than the per card belief that considers the precise card identity, the semantic belief considers the general actionability of a card (whether it is playable, discardable, or otherwise). Figure \ref{fig:per-card-semantic-scatter} illustrates the differences in correlation between cross-play and cross entropy between the two types of beliefs. Interestingly, we see that for SAD policies, the trend is more concrete for the semantic belief than the per card belief, suggesting that the conventions SAD policies form resolve how a card may be used relative to maximizing Dec-POMDP reward, rather than the exact card identity. In contrast, we can see that OP policies do in fact resolve per card identity. In this way we have demonstrated how our generalized belief models may be used to elucidate the blackbox of agent play (more applications in explainability may be found in Section \ref{sec:belief-introspec}). We speculate that for both types of beliefs, that training on a policy with a sufficiently high cross-play score with the test policy, or training with sufficiently many policies at training time, tends to give better generalization; since one in practice cannot control the former, we advocate the latter.

By training over a pool of policies, the generalized belief model may observe the different ways agents maintain symmetries in the Dec-POMDP, thus automating the learning of symmetries so to bypass strong environmental assumptions.

\section{Improving Cross-Play}\label{sec:improving-xp}

This section demonstrates how the generalized belief model may be leveraged to realise a policy suited for coordination.

\subsection{Method}\label{sec:xp-setup}

We propose two different schemes for leveraging the generalized belief: 1) a Monte Carlo search scheme; 2) training a best response over a pool of policies with the hidden state of the generalized belief model.

Search has been a critical component for the success of AI in several multi-agent settings, such as chess \cite{campbell2002deep}, poker \cite{moravvcik2017deepstack}, and Go \cite{silver2018general}. \citet{lerer2020improving} propose a method of note for the Dec-POMDP setting; their work maintains an exact belief using updates as in Equation (\ref{eq:dec-pomdp-belief-update}), samples from this belief to obtain hypothetical values for the unobservable state features, then conducts Monte Carlo search rollouts for each potential action predicated on these sampled state features using the agreed-upon policy to determine the best action. Motivated by this setup, we propose a similar framework for zero-shot cooperative tasks between policies $\pi^1, \dots, \pi^n$: we replace the exact belief with our learned generalized belief (thus bypassing costly Bayesian updating), and we conduct Monte Carlo rollouts by partnering $\pi^1$ with randomly chosen policies from the pool of policies the generalized belief was trained over. Intuitively, the generalized belief samples help infer the intent of policies $\pi^2$ through $\pi^n$, and the Monte Carlo rollouts calculate the most robust action for $\pi^1$ to take (by playing over rollouts of multiple different pool policies) predicated on this inference.

Along with the search scheme described above, we also consider training a best response. In general, one might not be able to expect high-level zero-shot performance between a best response trained over a pool of policies and a novel test policy. 
This is because, as mentioned in Section \ref{sec:beliefs}, it is difficult for the RL signal alone to form a rich belief representation \citep{moreno2018neural, gangwani2020learning}, especially in this setting over a pool of diverse policies. 
Furthermore, there is no guarantee that the consequent simple heuristics learned would generalize effectively to novel partners. Motivated by this, we propose training the best response with the hidden state of the generalized belief model. In this way, we essentially form a deeper, model-based architecture, and allow our best response to incorporate the demonstrated adaptability of the generalized belief to reason over novel conventions. In addition, training with the belief model's hidden state may even guide the formation of higher order beliefs (e.g., beliefs over beliefs). We train the best response using SAD and Independent Q-Learning \cite{tan1993multi}.
\looseness=-1

Refer to Appendix \ref{appendix:improving-xp} for more details on the search and best response setups.

\subsection{Results}\label{sec:xp-results}

\begin{table}[h]
% \vskip 0.15in
\begin{center}
\begin{scriptsize}
\begin{sc}
\begin{tabular}{lcccr}
\toprule
\textbf{SAD} & W/o & SBS & GBS\\
%Single    & 12.16$\pm$0.49 & 13.52$\pm$0.58& 13.81$\pm$ 0.56 \\
BR w/o gen. belief & 10.29$\pm$ 1.05& 11.32$\pm$ 1.18& 12.01$\pm$1.03\\
BR w/ gen. belief & 12.36$\pm$0.96 & 12.03$\pm$1.11 & 12.47$\pm$1.02\\
\bottomrule
\end{tabular}
\begin{tabular}{lcccr}
\toprule
\textbf{OP} & W/o & SBS & GBS\\
%Single    & 12.16$\pm$0.49 & 13.52$\pm$0.58& 13.81$\pm$ 0.56 \\
BR w/o gen. belief & 17.49$\pm$ 0.89& 17.81$\pm$ 0.92& 18.31$\pm$0.85\\
BR w/ gen. belief & 18.30$\pm$0.84 & 17.99$\pm$0.89 & 18.41$\pm$0.82\\
\bottomrule
\end{tabular}
\begin{tabular}{lcccr}
\toprule
\textbf{SAD for OP} & W/o & SBS & GBS\\
%Single    & 12.16$\pm$0.49 & 13.52$\pm$0.58& 13.81$\pm$ 0.56 \\
BR w/o gen. belief & 17.49$\pm$ 0.89& 18.47$\pm$ 0.85& 18.54$\pm$0.81\\
BR w/ gen. belief & 18.11$\pm$0.82 & 18.68$\pm$0.87 & 18.99$\pm$0.84\\
\bottomrule
\end{tabular}
\end{sc}
\end{scriptsize}
\end{center}
% \vskip -0.1in
\caption{Scores achieved by the various methods when playing against several test policies. The mean and standard error of the mean are reported here. W/o indicates the method without search applied on top. SBS (single belief search) is search applied with a Single belief model, and the quantity here denotes the average SBS across several Single beliefs. GBS (generalized belief search) is search applied with a generalized (Multi) belief model. Search considers 200 rollouts per legal move on the agent's turn.}
\label{table:xp-results}
\end{table}

Table \ref{table:xp-results} contains the results of our experiments on the generalized belief's ability to improve cross-play. We again use the thirteen SAD policies from \citet{hu2020other}, which achieve average cross-play scores of $6.258 \pm 1.51$, and twelve OP policies, which achieve average cross-play scores of $16.88 \pm 0.96$ (recall that $25$ is a perfect score in Hanabi). For each pool of policies, we randomly choose six of the polices to train a generalized belief model. We also train best response functions over these six policies (with and without the generalized belief hidden state as input), and use these six policies to conduct search rollouts as described in Section \ref{sec:xp-setup}. We then train six Single SAD models and six Single OP models using the six chosen policies from each pool, so to use these Single beliefs for search and compare performance with the generalized belief. We reserve the remaining policies ($13-6=7$ SAD policies, and $12-6=6$ OP policies) for testing to evaluate generalization ability. Of particular note, we test the ability of the SAD beliefs to improve cross-play of OP policies, thus testing generalization ability to policies outside of the population trained with.

To determine statistical significance between the methods in Table \ref{table:xp-results}, we conduct Monte Carlo permutation tests \cite{eden1933validity, dwass1957modified}, for which we bound each derived $p$-value in a 99\% binomial confidence interval.

We write the ``BR w/ gen. belief'' to refer to the best response trained with the generalized belief's hidden state as, and similarly write ``BR w/o gen. belief'' to refer to the best response trained without the generalized belief's hidden state. We write ``x with GBS applied'' and ``x with SBS applied'' to refer to a policy x with search applied, with the generalized belief and the single belief, respectively.

% \begin{figure}
%     \centering
%     \begin{minipage}{1.2\linewidth}
%     \hspace{-0.8cm}\includegraphics[width=1.0\linewidth]{prob-densities-methods.png}
%     \end{minipage}
%     \vspace{-1.0\baselineskip}
%     \caption{Probability densities of the scores achieved by the various methods with the seven test policies. Densities are ordered from left to right, from the highest scoring policy (in the average) to the least.}
%     \label{fig:prob-densities}
% \end{figure}

\begin{figure*}[ht!]
    \centering
    \includegraphics[width=0.8\linewidth]{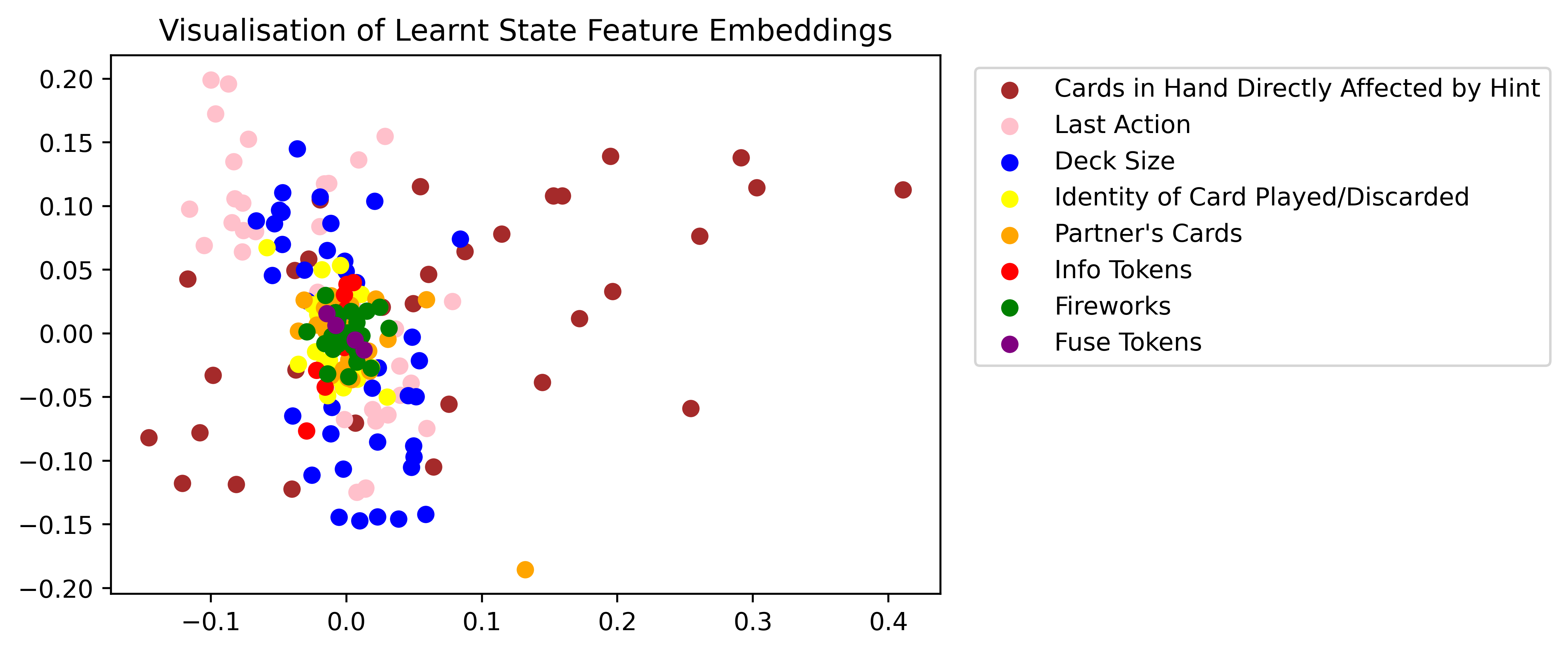}
    \caption{2-D projection of the embedding representations learned through the context of Hanabi gameplay of SAD policies (see Section \ref{sec:model} for how these representations are learned). The legend is ordered by features with most variance to least. More variance indicates more importance.}
    \label{fig:embedding-visual}
\end{figure*}

We first consider the intra-group cases; that is, using the SAD beliefs for SAD cross-play and OP beliefs for OP cross-play. The $p$-value for the one-tailed test for whether the SAD BR w/ gen. belief is different from the SAD BR w/o gen. belief is between $(0, 0.0000515]$, making the difference significant at the level $\alpha = 0.01$. For the OP case, the $p$-value is between $[0.0578, 0.0633]$, making the difference significant at the level $\alpha = 0.1$. Thus while both best responses trained with and without the generalized belief make use of the same training policies, the generalized belief offers a highly statistically significant edge for coordination for both pools of policies.

It can be seen that both SBS and GBS improve the performance of the BR w/o gen. belief for both SAD and OP. However, the $p$-value for the one-tailed test for whether the SAD BR w/o gen. belief with GBS applied is different from the SAD BR w/o gen. belief with SBS applied is between $[0.0808, 0.0872],$ making the difference significant at the level $\alpha=0.1$. In the OP case, the $p$-value is between $[0.0349, 0.0393]$, making the difference significant at the level $\alpha = 0.05$. As well, in the cases with the BR w/ gen. belief, SBS on average worsens performance. These observations in combination demonstrate the necessity of the generalized belief for search, and isolate its contribution from the search itself.

%The $p$-value for the one-tailed test for whether the BR w/o gen. belief with GBS applied is different from the BR w/o gen. belief \textit{without} search applied is between $(0, 0.000017]$, making the difference significant at the level $\alpha = 0.01$. The $p$-value drawn between the analogous one-tailed test comparing the BR w/o gen. belief with GBS applied and the BR w/o gen. belief with SBS applied is between $[0.0404, 0.0415]$, making the difference significant at the level $\alpha = 0.05$. Together these hypothesis tests show how GBS provides a highly statistically significant edge for the agent, and that the generalized belief is moreover still statistically significantly better than the Single belief, in spite of the search procedures using a number of different policies to conduct rollouts.

The one-tailed test between the SAD BR w/o gen. belief with GBS applied and the SAD BR w/o gen. belief without search applied finds the $p$-value to be between $[0.00131, 0.00223]$, making the difference significant at the level $\alpha = 0.01$. In contrast, the one-tailed test between the BR w/ gen. belief with GBS applied and the BR w/ gen. belief without search applied finds the $p$-value to be between $[0.451, 0.462]$, and similarly for the OP case the $p$-value is between $[0.393, 0.404]$. This shows that further applying GBS onto the BR w/ gen. belief does not confer a statistically significant advantage, in spite of search averaging over thousands of variations with rollouts leveraging a diverse pool of agents, and as such suggests that the BR w/ gen. belief has learned sufficiently high-order play to bypass the need for costly search rollouts.

Now we consider the inter-group case;, that is, using the SAD beliefs for OP cross-play. The $p$-value for the one-tailed test for whether the BR w/ gen. belief is different from the BR w/o gen. belief is between $[0.0431, 0.0467],$ making the difference significant at the level $\alpha = 0.05$. The $p$-value for the one-tailed test for whether the BR w/ gen. belief with GBS applied is different from the BR w/ gen. belief with SBS applied is between $[0.0926, 0.0978]$, making the difference significant at the level $\alpha = 0.1$. This shows that improvement can be realised across policies from different populations (e.g. SAD beliefs for OP policies), evincing the generality of our method.

Sections \ref{sec:generalized-beliefs} and \ref{sec:improving-xp} demonstrate in combination over an array of policies and over a variety of methods that the generalized belief indeed learns generalizable conventions, such that in Section \ref{sec:improving-xp} the generalized belief's adaptability yields improved cross-play. Further, the generalized belief allows improvement in cross-play without having to constrain against the conventions the policies are able to form.
Next, we take a closer look at what the belief model learns.

\section{Belief Model Introspection}\label{sec:belief-introspec}

\paragraph{Attention Heads}

One of the advantages of attention-based architectures is the transparency and interpretability of the attention weights. Appendix \ref{appendix:attention-heads} shows some of the attention patterns learned in our autoregressive model, that show how ``pertinent'' the model views a given timestep $x$ when considering a timestep $y$.

\paragraph{State Feature Embeddings}

%comment{\begin{figure*}[H]
%    \centering
%    \includegraphics[width=1.0\linewidth]{visual-embeddings.png}
%    \caption{2-D projection of the embedding representations learned through the context of Hanabi gameplay (see Section \ref{sec:model} for how these representations are learned). Legend is ordered by features with most variance to least. More variance indicates more importance.}
%    \label{fig:embedding-visual}
%\end{figure*}}

We can also visualise the representations learned for the observable state features by applying singular value decomposition \cite{eckart1936approximation} to the $K \times d_\text{feature}$ matrix of feature embeddings (see Section \ref{sec:model}) and considering the first two singular values to obtain the principal components as in Figure \ref{fig:embedding-visual}.

Because we input the state features into $\Psi_E$ in a fixed concatenated order, the linear layer that aggregates the projected representations of the features learns this order and weights each feature accordingly. Hence for our context, the position each state feature embedding occupies in the vector space relative to different types of state features does not matter. However, the distribution of principal components relative to a single state feature type, and in particular, which state feature types have the most variance, indicates the expressiveness of a state feature type (if all the learned representations of a state feature type are similar, then the state feature is not very informative). 

Figure \ref{fig:embedding-visual} shows that the observable state feature types with the most variation are which cards in hand were directly hinted,\footnote{We say ``directly'' hinted to mean the cards that receive the explicit grounded information: e.g. if I hint that your second and fourth cards are of rank 1, then your second and fourth cards were directly hinted; meanwhile your first, third and fifth cards were indirectly hinted of not being of rank 1.} the last action executed in the game, and the remaining size of the deck, indicating that the model found these features to be most informative for forming belief. While it is clear that much of the grounded information is manifested in these state features, we conclude that the policy also must mostly use these features to convey the implicit information because the variances of the other features are much lower. This may take the form of a secondary meaning attached to a hint based on which cards were directly hinted, or a pre-agreed convention on how to act relative to the number of cards remaining in the deck.

Thus while it may in general be difficult to interpret the nuanced tactics employed by an expert policy, these visual analytics can provide evidence regarding the style of play.

\section{Related Work}

\paragraph{Learning to Coordinate}
A central challenge in AI is devising agents that can coordinate with a novel partner \cite{canaan2020evaluating, carroll2019utility, hu2020other, kleiman2016coordinate, papoudakis2020variational, smith2020learning, stone2010ad,  zintgraf2021deep}. As mentioned in Section \ref{sec:intro}, naively applying self-play approaches to one another can yield poor performance. While dynamic programming approaches may be considered as an approach to Dec-POMDPs \cite{hansen2004dynamic}, such planning methods assume that at test time an agent's teammates will execute their part of the same centrally planned joint policy as them. In \citet{hu2020other}, a policy is taught to pay attention to symmetry in the Dec-POMDP at training time so as to form generalizable conventions that can enhance cross-play performance. In this way, the agent aims to learn an unambiguous but potentially sub-optimal policy. In \citet{shih2021critical}, a modular approach is adopted to separately learn rule-dependent and convention-dependent behaviours to facilitate coordination, where their experiments are focused on a small version of Hanabi. See Section \ref{sec:intro} for how our method compares with \citet{hu2020other} and \citet{shih2021critical}.
% \looseness=-1

\citet{lupu2021trajectory} consider training a diverse pool of policies suited for zero-shot coordination, whereas we focus on aggregating the conventions of existing policies to generalize over new ones. \citet{hu2021off} propose using a belief model trained over a fixed policy, and train a new policy that predicates decisions on samples from the trained belief. The ability of the new policy to then effectively coordinate with a test class of partner policies is highly dependent on the relation between the fixed policy the belief is trained on, and the test policies. In contrast, we train our autoregressive belief on multiple policies to aggregate a diversity of convention-dependent behaviours. 

An important difference with our work and other approaches is as follows: existing literature often proposes methods for constraining policy training in such a way to allow trained policies to effectively engage in cross-play amongst one another, and these works focus experimentation in this realm. In contrast, we propose a method for improving cross-play ability against classes of policies not necessarily pre-configured for cross-play (i.e. ad-hoc teamplay), and demonstrate that this can indeed be done.

\paragraph{Self-Attention and Transformers in Reinforcement Learning}
\citet{parisotto2020stabilizing} proposed the addition of extra layers into the transformer architecture to stabilize training in the high-variance RL domain, where these additional layers could possibly be used to enhance the methodology proposed in this paper, which we leave for future work. \citet{chen2021decision} proposed using the transformer architecture for tasks in offline RL. \citet{hu2021updet} explore the usage of semantic embeddings over observable features for the transformer architecture, but we improve both on input scalability and providing context to the data by aggregating the features with a fully-connected layer (see Section \ref{sec:model}).

\section{Conclusion}

In this paper, we proposed a method for learning a generalized belief across multiple policies, where the model learns to reason over the specialized conventions of each policy. We tested this model on diverse collections of policies that, such that especially for the case of the SAD policies, were unable to collaborate well in cross-play due to their high degree of conventional specialization. Our proposed model was nonetheless able to factorise the respective conventions of these policies and understand the implicit information they conveyed. We also showed that training over a set of conventions can suffice for generalizing to new ones, so that one can decode the intent of a given trajectory based on the known intents of other policies and the learned symmetry in the Dec-POMDP; we showed all this can be done without strong environmental assumptions, and without constraining policy training. We further leveraged this belief model to improve coordination, and proposed two frameworks under which the generalized belief can be used. In particular, we found that inter-population generalization was possible with our frameworks, such that beliefs trained over one population could be used to improve coordination over another (i.e. SAD beliefs for OP policies).

In addition, we proposed architectural adaptations to suit the transformer (and other self-attention based models) to the MARL domain, and empirically verified the ability for a model to effectively learn under the architectural adaptations. Finally, we proposed visual analytic schemes to help with interpretability of the belief and policy, which is critical for increasing transparency and trust of RL systems \cite{wells2021explainable}.

In future work, we will explore even larger Dec-POMDPs, perhaps larger variants of Hanabi, or realistic settings: for realistic settings, one has a choice in one of two beliefs: 1) enumerating the unobservable features of the environment and learning the per-feature belief; 2) considering only the agent's action space and learning the semantic belief. One may train a best response with either belief, as in the paper, or with the per-feature belief one may additionally use a search scheme, as in the paper.

\section*{Acknowledgements}

The experiments were made possible by a generous equipment grant from NVIDIA. Luisa Zintgraf is supported by the 2017 Microsoft Research PhD Scholarship Program, and the 2020 Microsoft Research EMEA PhD Award. Christian Schroeder de Witt is generously funded by the Cooperative AI Foundation.

% % In the unusual situation where you want a paper to appear in the
% % references without citing it in the main text, use \nocite
\bibliography{example_paper}
\bibliographystyle{icml2022}

%%%%%%%%%%%%%%%%%%%%%%%%%%%%%%%%%%%%%%%%%%%%%%%%%%%%%%%%%%%%%%%%%%%%%%%%%%%%%%%
%%%%%%%%%%%%%%%%%%%%%%%%%%%%%%%%%%%%%%%%%%%%%%%%%%%%%%%%%%%%%%%%%%%%%%%%%%%%%%%
% APPENDIX
%%%%%%%%%%%%%%%%%%%%%%%%%%%%%%%%%%%%%%%%%%%%%%%%%%%%%%%%%%%%%%%%%%%%%%%%%%%%%%%
%%%%%%%%%%%%%%%%%%%%%%%%%%%%%%%%%%%%%%%%%%%%%%%%%%%%%%%%%%%%%%%%%%%%%%%%%%%%%%%
\newpage
\appendix
\onecolumn

\section{Hanabi}\label{appendix:hanabi-rules}
Hanabi is a cooperative card game that can be played with 2 to 5 people. Hanabi is a popular game, having been crowned the 2013 ``Spiel des Jahres'' award, a German industry award given to the best board game of the year. Hanabi has been proposed as an AI benchmark task to test models of cooperative play that act under partial information \cite{bard2020hanabi}. To date, Hanabi has one of the largest state spaces of all Dec-POMDP benchmarks.

The deck of cards in Hanabi is comprised of five colours (white, yellow, green, blue and red), and five ranks (1 through 5), where for each colour there are three 1's, two each of 2's, 3's and 4's, and one 5, for a total deck size of fifty cards. Each player is dealt five cards (or four cards if there are 4 or 5 players). At the start, the players collectively have eight information tokens and three fuse tokens, the uses of which shall be explained presently.

In Hanabi, players can see all other players' hands but their own. The goal of the game is to play cards so as to collectively form five consecutively ordered stacks, one for each colour, beginning with a card of rank 1 and ending with a card of rank 5. These stacks are referred to as fireworks, as playing the cards in order is meant to draw analogy to setting up a firework display.\footnote{Hanabi (\begin{CJK*}{UTF8}{gbsn}花火\end{CJK*}) means `fireworks' in Japanese.}

We call the player whose turn it is the active agent. The active agent must conduct one of three actions:

\begin{itemize}
    \item \textbf{Hint} - The active agent chooses another player to grant a hint to. A hint involves the active agent choosing a colour or rank, and revealing to their chosen partner all cards in the partner's hand that satisfy the chosen colour or rank. Performing a hint exhausts an information token. If the players have no information tokens, a hint may not be conducted and the active agent must either conduct a discard or a play.
    
    \item \textbf{Discard} - The active agent chooses one of the cards in their hand to discard. The identity of the discarded card is revealed to the active agent and becomes public information. Discarding a card replenishes an information token should the players have less than eight.
    
    \item \textbf{Play} - The active agent attempts to play one of the cards in their hand. The identity of the played card is revealed to the active agent and becomes public information. The active agent has played successfully if their played card is the next in the firework of its colour to be played, and the played card is then added to the sequence. If a firework is completed, the players receive a new information token should they have less than eight. If the player is unsuccessful, the card is discarded, without replenishment of an information token, and the players lose a fuse token.
    
\end{itemize}

The game ends when all three fuse tokens are spent, when the players successfully complete all five fireworks, or when the last card in the deck is drawn and all players take one last turn. If the game finishes by depletion of all fuse tokens, the players receive a score of 0. Otherwise, the score of the finished game is the sum of the highest card ranks in each firework, for a highest possible score of 25.

\newpage

\section{Autoregressive Beliefs}\label{appendix:autoregressive-beliefs}

Firstly, we assume that we can factor the unobservable state features from $s_t$, as is the usual case for Dec-POMDPs. Proceeding, let $H = |\{c_1, \dots, c_H\}|$ be the number of unobservable state features, and assume $C$ is the number of different values each $c_h$ may take on. We thus take the belief to be over these unobservable features, and consider the identification of belief learning as an autoregressive task \citep{hu2021learned}; namely, to find
\begin{equation}\label{eq:belief-learning-as-seq2seq}
    \psi ~~ \text{ s.t. } ~~ c_h \sim \psi(c_h \ | \ \textbf{c}_{<h}, \tau^i_t),
\end{equation}
where $\psi$ is our probabilistic mapping between the action-observation history $\tau^i_t$ and the unobservable state features $c_1,\dots,c_H$, where $\textbf{c}_h := (c_1, \dots, c_{h-1})$ and $\textbf{c}_1 := \emptyset$. In which case, for $\psi$ parameterized as $\psi_\theta$, we produce the belief approximation
\begin{align*}
    b_t^i \approx \prod_{h=1}^H\psi_\theta(c_h \ | \ \textbf{c}_{<h}, \tau_t^i), \numberthis
\end{align*}
which we learn in a supervised fashion through minimization of the cross entropy loss,
\begin{align*}
     CE(\psi_\theta, \tau_t) = -\frac{1}{H}\sum_{h=1}^H\sum_{x=1}^C \mathds{1}_{x=c_h}(x \ | \ s_t) \log \psi_\theta(x \ | \ \textbf{c}_{<h}, \tau_t^i).\numberthis
\end{align*}

\section{Cross Entropy Interpretation}\label{appendix:ce-interpretation}

\begin{definition}\label{def:narrow-down}
    For an agent $i$, $\psi$ is said to have narrowed down $c_1,\dots,c_H$ over $\tau_t^i$ to at most $n$ if $n\geq e^{\mathbb{E}_{p(\tau_t)} [CE(\psi, \tau_t)]}$.
\end{definition}

This definition is motivated as follows: first suppose the indicator function corresponding to each unobservable state feature $h$ is chosen uniformly at random over $C$. Then,
\begin{align*}
    \mathbb{E}_{p(\mathds{1})} [CE(\psi, \tau_t)] &= -\frac{1}{CH}\sum_{h=1}^H\sum_{x=1}^C\log \psi(x \ | \ \textbf{c}_{<h}, \tau_t^i)
    \\&\geq -\log \big(\frac{1}{CH}\sum_{h=1}^H\sum_{x=1}^C \psi(x \ | \ \textbf{c}_{<h}, \tau_t^i)\big)
    \\&=-\log \big(\frac{1}{CH}\sum_{h=1}^H1\big)
    \\&= -\log\frac{1}{C},
\end{align*}
where the second line follows from Jensen's inequality and convexity of the negative logarithm \cite{jensen1906fonctions}, and the third line follows from probabilities summing to $1$. 

But note that in a Dec-POMDP where the distribution over unobservable state features is uniform (e.g. a shuffled deck of cards in Hanabi), we have
\begin{equation}
  \mathbb{E}_{p(\mathds{1})} [CE(\psi, \tau_t)] = \mathbb{E}_{p(\tau_t)} [CE(\psi, \tau_t)].  
\end{equation}
Hence $n \geq e^{\mathbb{E}_{p(\tau_t)} [CE(\psi, \tau_t)]} \geq e^{-\log\frac{1}{C}} = C$; that is, the number of different values the unobservable features may take on is at most $n$.

One can thus use the Monte Carlo estimate
\begin{equation}
    e^{\mathbb{E}_{p(\tau_t)} [CE(\psi, \tau_t)]} \approx e^{\frac{1}{m}\sum_{\tau_{t,l}} CE (\psi,\tau_{t,l})},
\end{equation}
where $\sum_{\tau_{t,l}}$ denotes a sum over $m$ randomly chosen trajectories.

To give examples, for the Hanabi control task, when the belief model achieves an average cross entropy score less than or equal to $-\log\frac{1}{5} \approx 1.609$, it will have narrowed down the number of possible cards that the average card in hand can be to at most $5$. When the model knows the exact identity of a card, it will achieve a cross entropy score of $-\log 1 = 0.$

\clearpage

\section{Experimental Setup}

We base our training setup on the codebases of \citet{hu2019simplified, hu2021off, hu2020other}, which we link here: \url{https://github.com/facebookresearch/hanabi_SAD}, \url{https://github.com/facebookresearch/off-belief-learning}. With these links, one will also find the pre-trained SAD policies used in this work.

The machine used for experimentation consisted of 2 NVIDIA GeForce RTX 2080 Ti GPUs and 40 CPU cores.

\subsection{Belief Learning}\label{appendix:belief-learning}

The code for learning a belief model with the transformer architecture may be found here: \url{https://github.com/gfppoy/hanabi-belief-transformer}.

\begin{table}[!htb]
    \centering
    \caption{Hyperparameter settings of transformer for belief emulation.}
    \begin{tabular}[t]{lcc}
    \toprule
    Hyperparameters&Value\\
    \midrule
    Number of layers&6\\
    Number of attention heads&8\\
    State embedding dimension ($d$ in Section \ref{sec:model})&256\\
    Feature embedding dimension ($d_\text{feature}$ in Section \ref{sec:model})&128\\
    Maximum sequence length ($T$ in Section \ref{sec:model})&80\\
    Feedforward network dimension&2048\\
    Nonlinearity&ReLU\\
    Batchsize&256\\
    Dropout&0.1\\
    Learning rate&2.5 $\times$ 10$^\text{-4}$\\
    Warm-up period&10$^\text{5}$\\
    Learning rate decay&Inverse square root\\
    \bottomrule
    \end{tabular}
\end{table}%

\subsection{Improving Cross-Play}\label{appendix:improving-xp}

The code for learning a best response without the belief model hidden state may be found here: \url{https://github.com/gfppoy/hanabi-br-wobelief}.

The code for learning a best response with the belief model hidden state may be found here: \url{https://github.com/gfppoy/hanabi-br-withbelief}.

The code for generalized belief search may be found here: \url{https://github.com/gfppoy/hanabi-gbs}.

\clearpage

\section{Attention Head Visualisations}\label{appendix:attention-heads}
The following are visualisations of the attention heads in our generalized belief models trained over the SAD policies, where here it is emulating the belief distribution at the $45^\text{th}$ timestep of a randomly generated game of Hanabi; a point $(x,y)$ on the visualisation represents how ``pertinent'' the model views timestep $x$ when considering timestep $y$, where $x$ varies along the horizontal axis and $y$ along the vertical axis. The visualisation shows the model found that considering  adjacent turns provides more context for deducing implicit cooperative cues.

\begin{figure}[hbt!]
    \centering
    \includegraphics[width=1.0\linewidth]{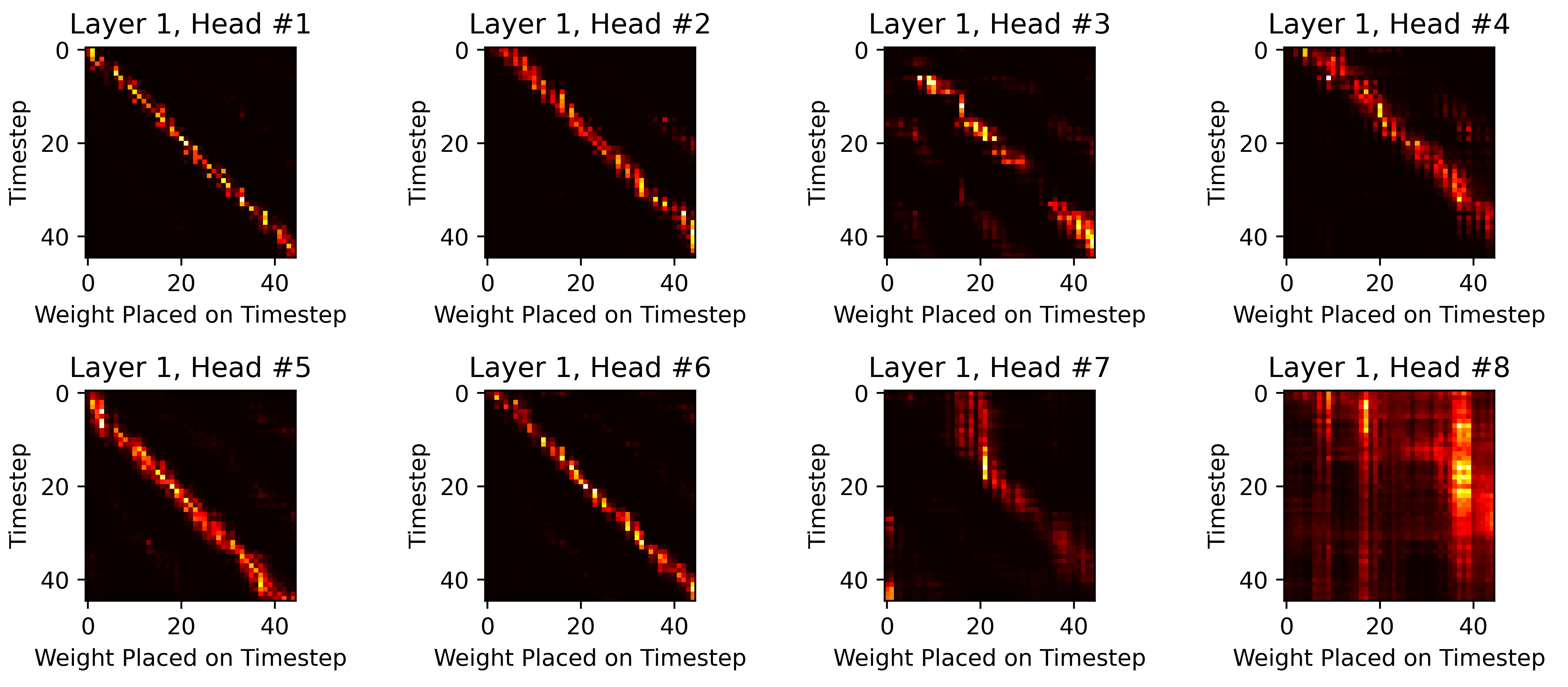}
    \vspace{-2.0\baselineskip}
    \caption{The $8$ attention head weights in the first layer.}
    %\label{fig:layer-1-attention}
\end{figure}
\begin{figure}[hbt!]
    \centering
    \includegraphics[width=1.0\linewidth]{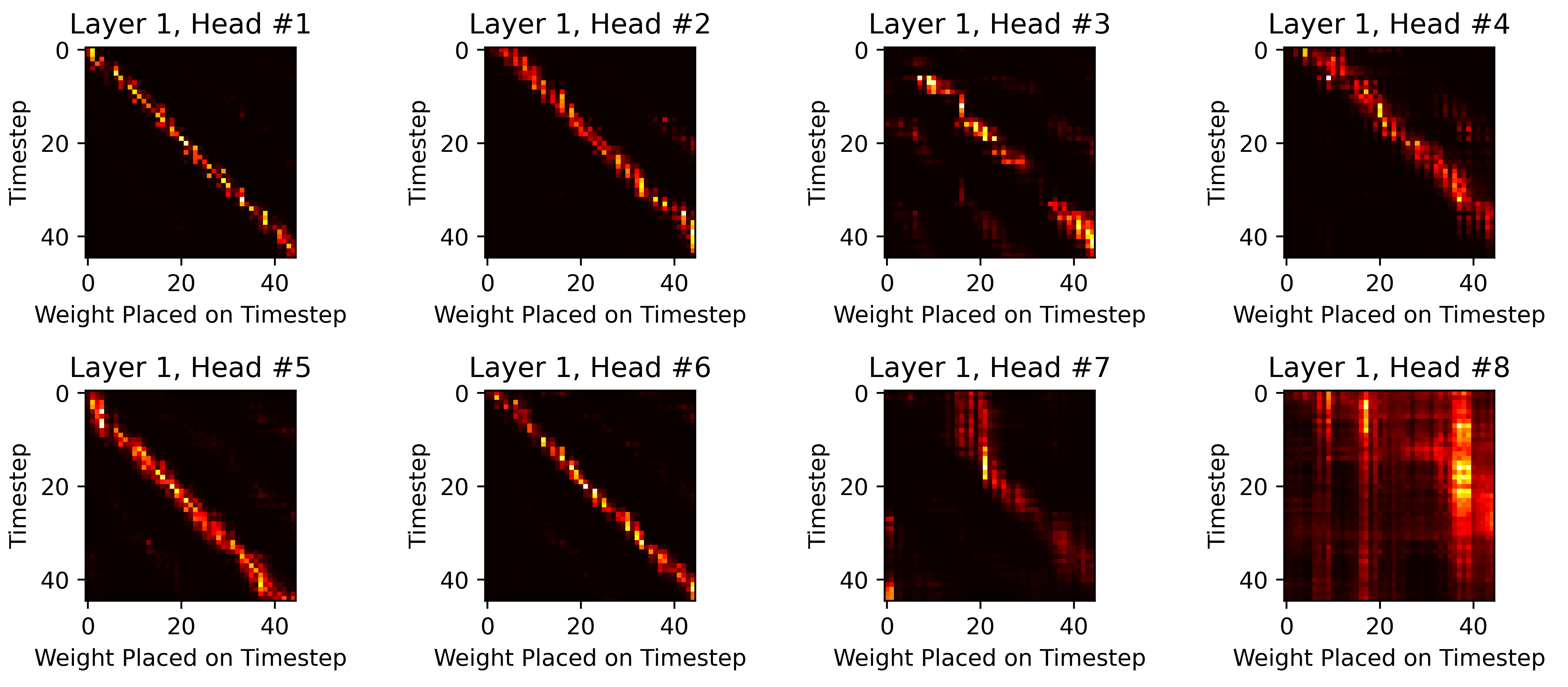}
    \caption{The $8$ attention head weights in the first layer.}
\end{figure}
\begin{figure}[hbt!]
    \centering
    \includegraphics[width=1.0\linewidth]{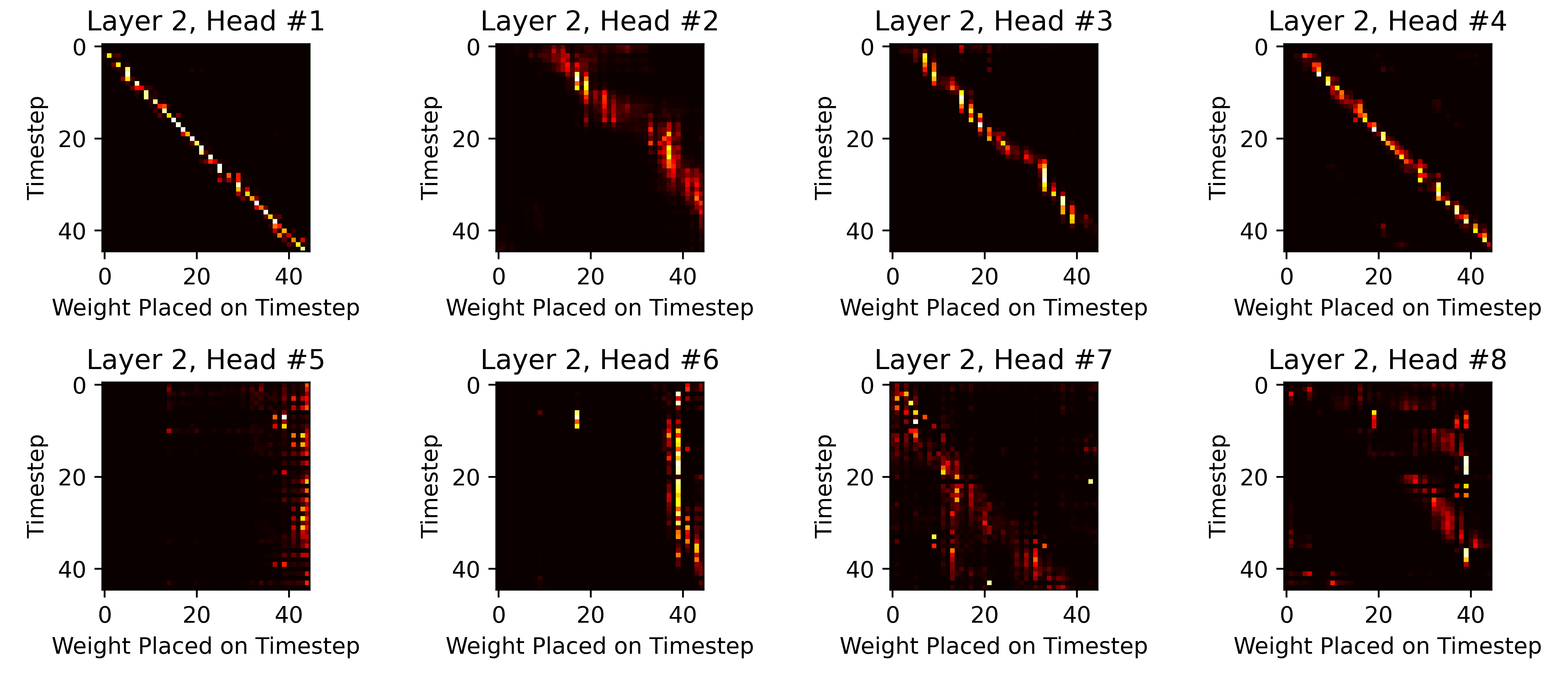}
    \caption{The $8$ attention head weights in the second layer.}
\end{figure}
\begin{figure}[hbt!]
    \centering
    \includegraphics[width=1.0\linewidth]{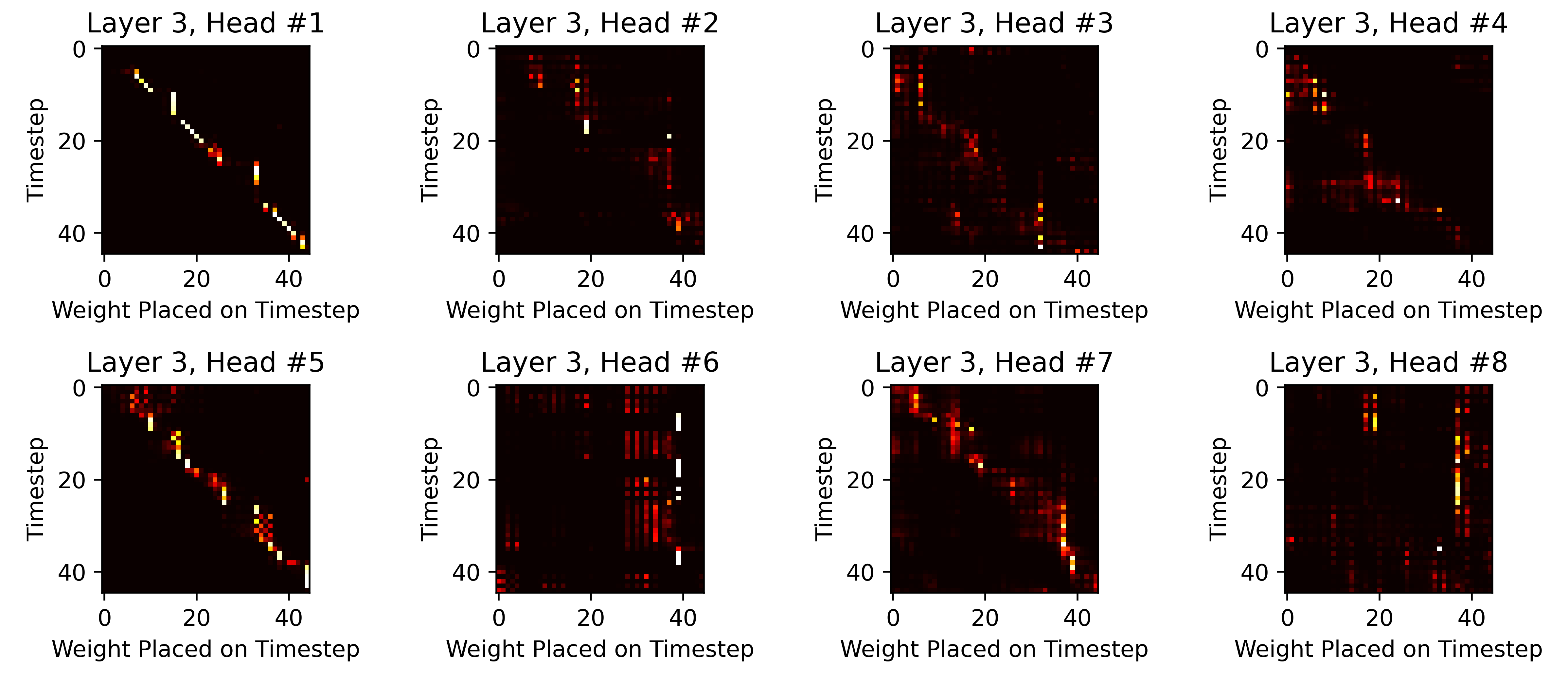}
    \caption{The $8$ attention head weights in the third layer.}
\end{figure}
\begin{figure}[hbt!]
    \centering
    \includegraphics[width=1.0\linewidth]{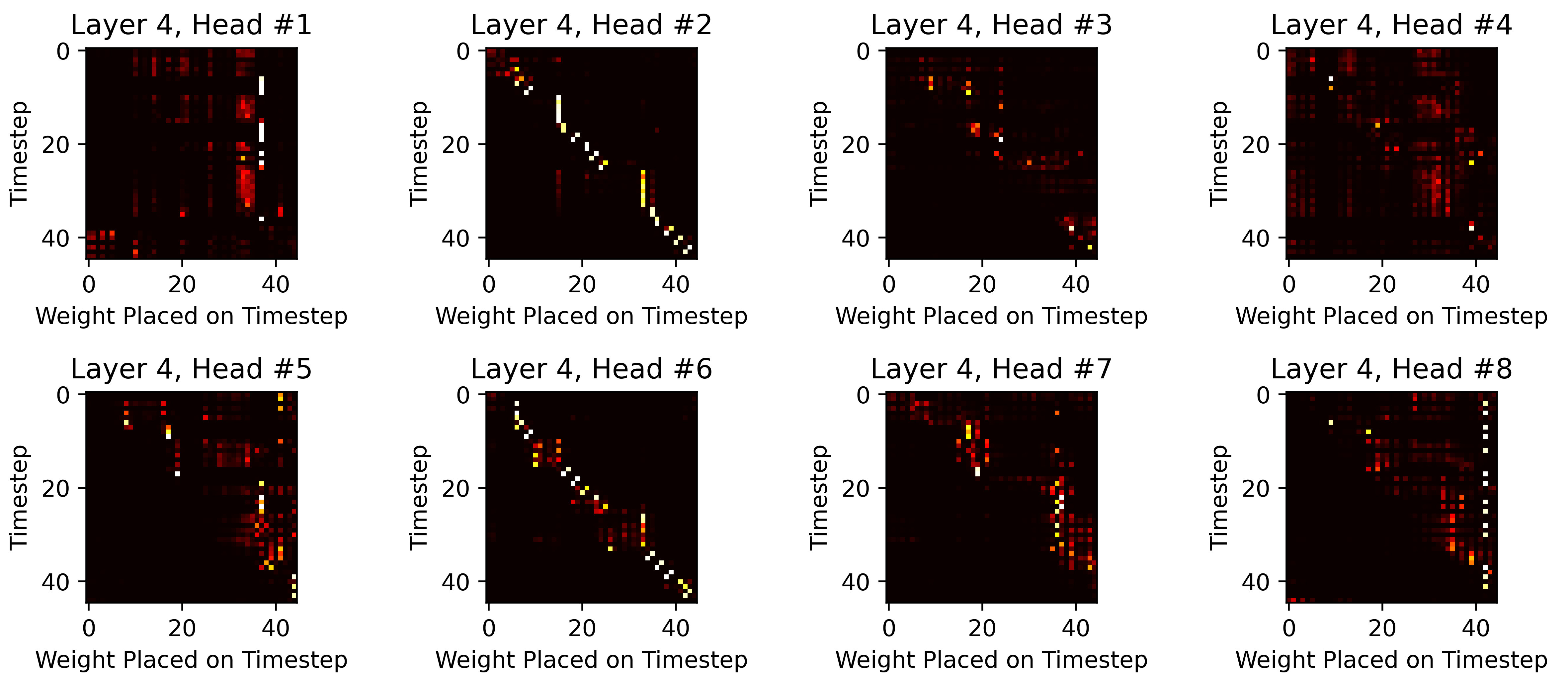}
    \caption{The $8$ attention head weights in the fourth layer.}
\end{figure}
\begin{figure}[hbt!]
    \centering
    \includegraphics[width=1.0\linewidth]{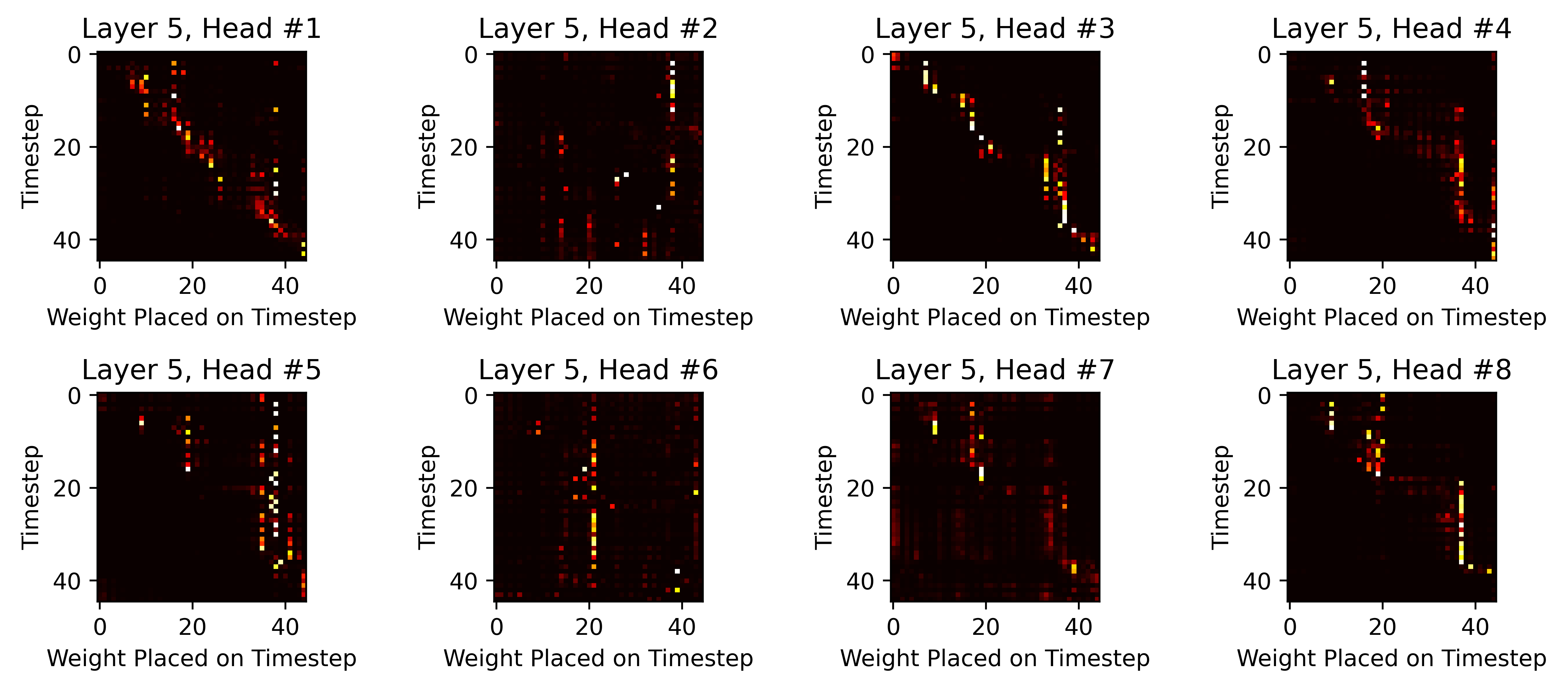}
    \caption{The $8$ attention head weights in the fifth layer.}
\end{figure}
\begin{figure}[hbt!]
    \centering
    \includegraphics[width=1.0\linewidth]{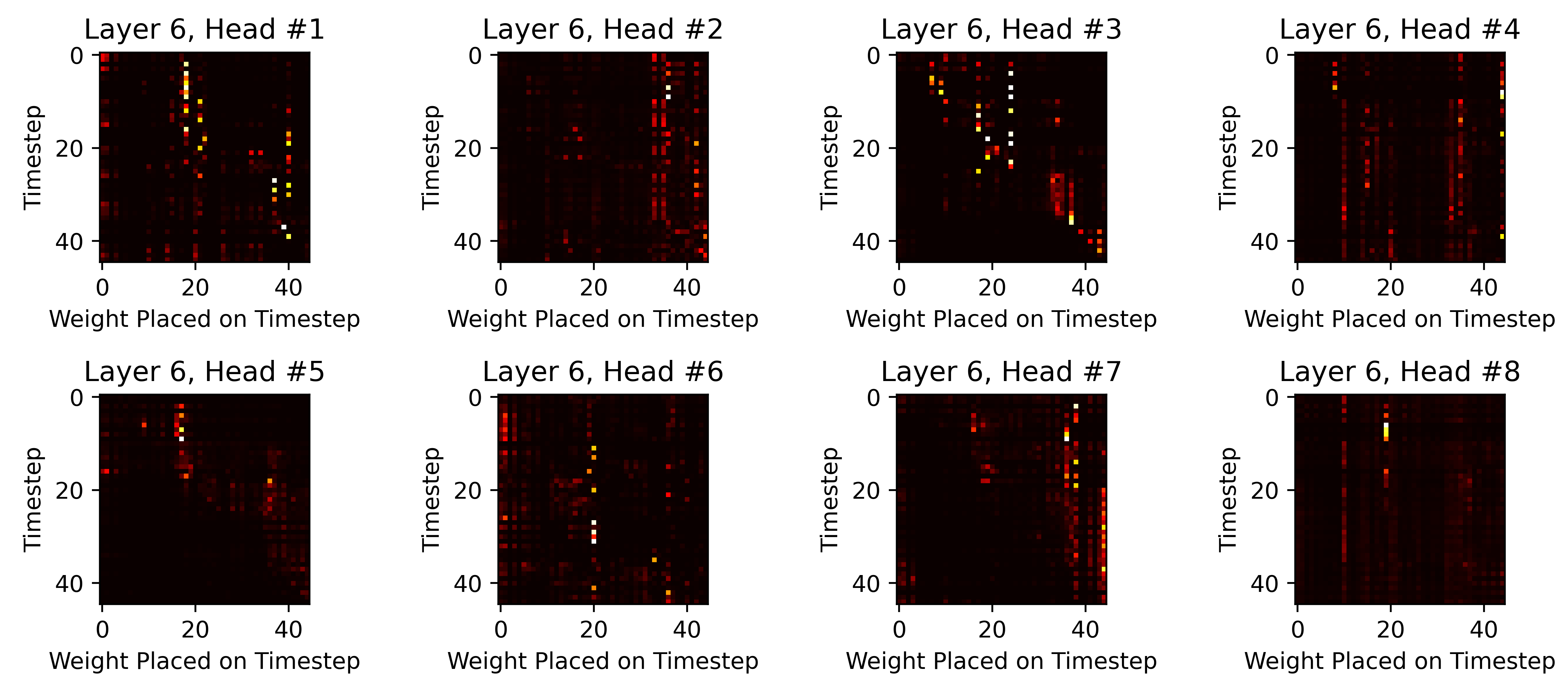}
    \caption{The $8$ attention head weights in the sixth layer.}
\end{figure}

\clearpage

\section{Distributions of Play}\label{appendix:play-distributions}
The figures illustrate the Hanabi cross-play score frequencies of various policies matched with SAD and OP policies, respectively. The best responses and beliefs here are trained over a pool of SAD policies and OP policies, respectively.
\begin{figure}[hbt!]
    \centering
    \includegraphics[width=1.0\linewidth]{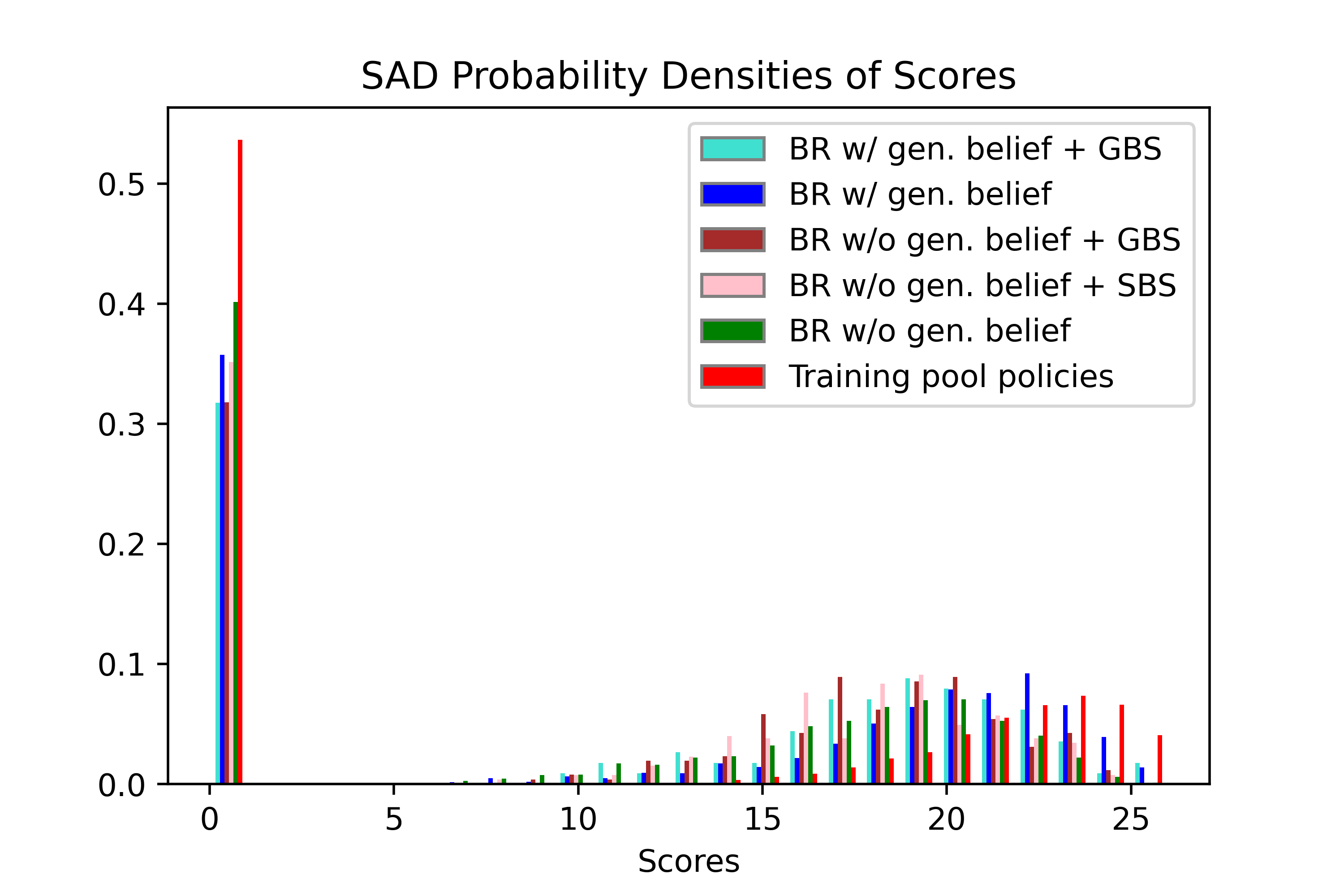}
    \caption{Probability densities of scores attained by various SAD models.}
\end{figure}
\begin{figure}[hbt!]
    \centering
    \includegraphics[width=1.0\linewidth]{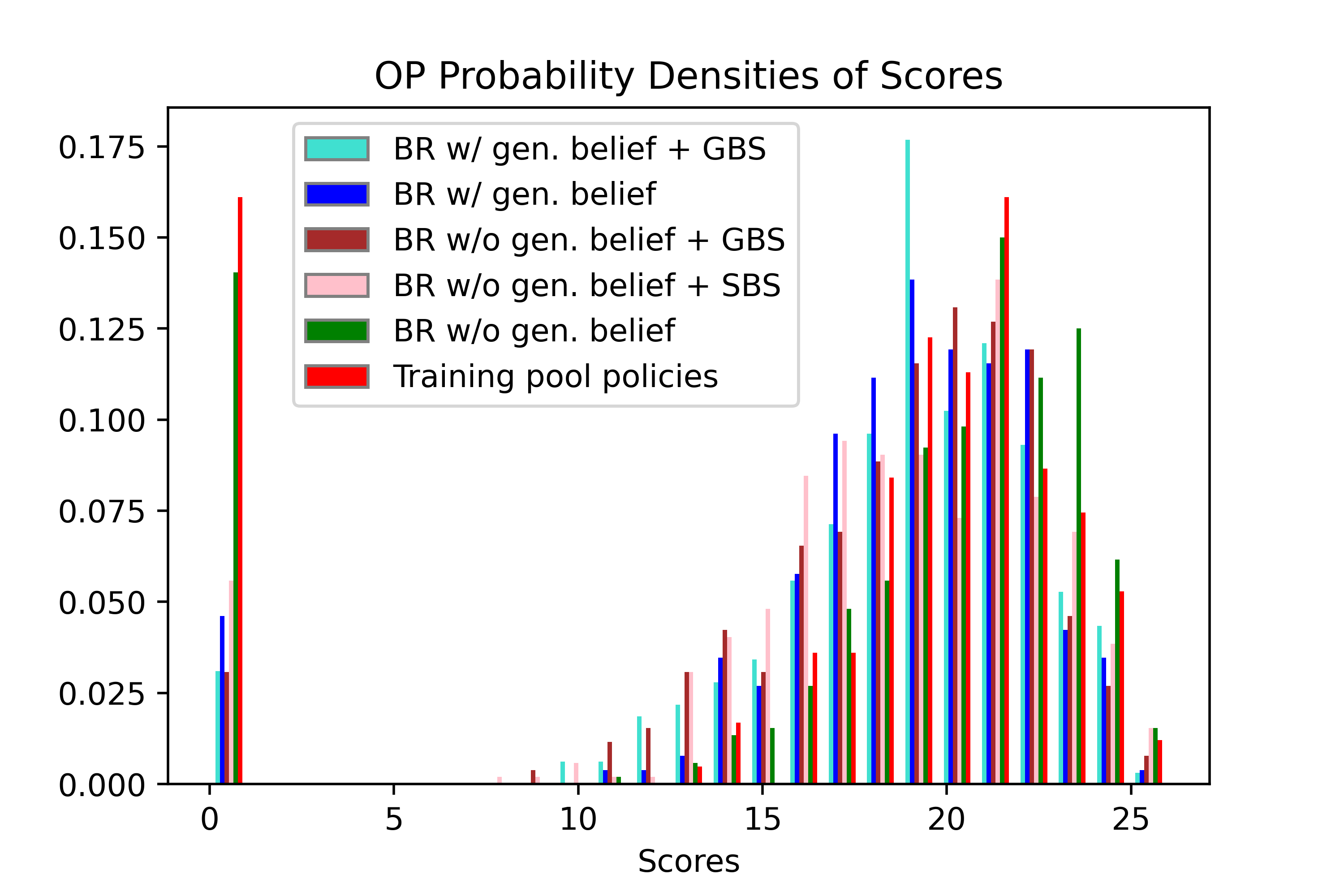}
    \caption{Probability densities of scores attained by various OP models.}
\end{figure}

\clearpage

\section{LSTM vs Transformer Beliefs}
Here we compare the performance of an LSTM encoder-decoder architecture with that of the transformer architecture used in this work for maintaining belief of policies not seen at training time. Transformers have advantages in parallelizability and interpretability, but here we demonstrate additional advantages in the ability to maintain belief. For both architectures, we use our proposed embedding mechanism (see Section \ref{sec:model}).
\begin{figure*}[hbt!]
    \centering
    \includegraphics[width=1.0\linewidth]{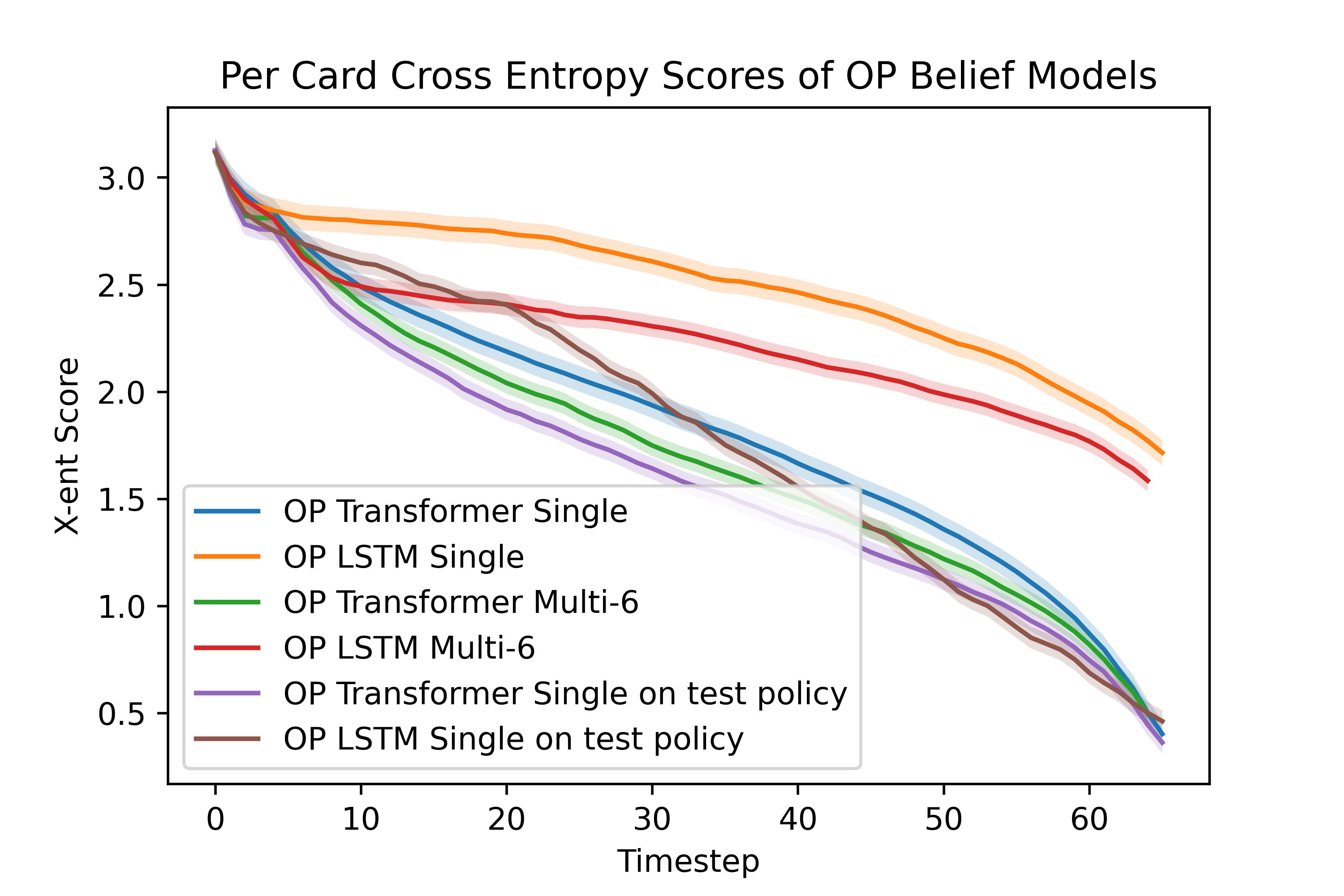}
    \caption{Per card cross entropy (X-ent) scores of the averages of the Single models and Multi-6 models, all tasked with maintaining belief over trajectories featuring a policy not seen at training time. The belief model trained on the test policy itself is provided here for reference. The shading corresponds to the standard error of the mean at each timestep. The curves were computed over 20k randomly generated games.}
    \label{fig:lstm-vs-transformer}
\end{figure*}

%%%%%%%%%%%%%%%%%%%%%%%%%%%%%%%%%%%%%%%%%%%%%%%%%%%%%%%%%%%%%%%%%%%%%%%%%%%%%%%
%%%%%%%%%%%%%%%%%%%%%%%%%%%%%%%%%%%%%%%%%%%%%%%%%%%%%%%%%%%%%%%%%%%%%%%%%%%%%%%

\end{document}